%% file: colm2024_conference.tex
\documentclass{article}
\usepackage[utf8]{inputenc}
\usepackage{colm2024_conference}

\usepackage{booktabs}
\usepackage{graphicx}
\usepackage{enumitem}
\usepackage{wrapfig}
\usepackage{algorithm}
\usepackage{algpseudocode}
\usepackage{natbib}
\usepackage{makecell}
\usepackage{booktabs}
\usepackage{array}
\usepackage{amsmath}
\usepackage{amssymb}
\usepackage{amsfonts}
\usepackage{multirow}
\usepackage{verbatim}
\usepackage{caption}
\usepackage{longtable}
\usepackage{supertabular}
\usepackage{CJKutf8}
\usepackage[utf8]{inputenc} 
\usepackage[T1]{fontenc}
\usepackage[french,vietnamese,mongolian,greek,english]{babel}
\usepackage{pifont}

\usepackage{enumitem}
\usepackage{tablefootnote}
\usepackage{xspace}
\usepackage{textcomp}
\usepackage{makecell}
\usepackage{lscape} 
\usepackage{siunitx}
\usepackage{listings}
\usepackage{xcolor}
\usepackage{svg}
\definecolor{dt}{gray}{0.7}
\usepackage{pifont}       
\usepackage{bbding}       
\usepackage{fontawesome}

\usepackage{scrextend}
\usepackage{ulem}
\usepackage{tgpagella}
\usepackage{latexsym}
\usepackage[T1]{fontenc}
\usepackage[utf8]{inputenc}
\usepackage{microtype}
\definecolor{mydarkblue}{rgb}{0,0.08,0.45}
\definecolor{citecolor}{HTML}{0071BC}
\usepackage{url}            
\usepackage{nicefrac}       
\usepackage{changepage}
\usepackage{xargs}          
\usepackage{wrapfig,lipsum,booktabs}
\usepackage{longtable}
\usepackage{subcaption}
\usepackage{endnotes}

\usepackage{pgfplots}
\usetikzlibrary{pgfplots.groupplots}
\pgfplotsset{compat=1.3}
\usepackage{tikz}
\usetikzlibrary{patterns}

\usepackage[most]{tcolorbox}

\usepackage{hyperref}
\definecolor{darkblue}{rgb}{0, 0, 0.5}
\hypersetup{colorlinks=true, citecolor=darkblue, linkcolor=red, urlcolor=darkblue, bookmarksopen=false, bookmarksnumbered=true}

\usepackage[capitalize,noabbrev]{cleveref}
\newcommand{\ours}{Tstars-Tryon 1.0\xspace}

\crefname{section}{\S}{\S\S}
\Crefname{section}{\S}{\S\S}
\crefname{subsection}{\S\S}{\S\S}
\Crefname{subsection}{\S\S}{\S\S}
\crefformat{section}{\S#2#1#3}
\crefformat{subsection}{\S\S#2#1#3}

\crefname{table}{Table}{Tables}
\crefname{figure}{Figure}{Figures}
\crefname{algorithm}{Algorithm}{}
\crefname{equation}{eq.}{}
\crefname{appendix}{Appendix}{}
\crefformat{section}{Section #2#1#3}
\usepackage{multicol}
\usepackage{tcolorbox}
\usepackage{titlesec}
\titleformat*{\section}{\large\bfseries}
\usepackage{array} 
\newcolumntype{P}[1]{>{\centering\arraybackslash}p{#1}} 
\usepackage{multirow}

\usepackage{adjustbox}
\definecolor{objblue}{RGB}{3,139,221}  
\definecolor{attrred}{RGB}{255,67,67}    
\definecolor{easygreen}{RGB}{0,156,75}  
\definecolor{middleyellow}{RGB}{242,89,34}  
\definecolor{hardred}{RGB}{216,56,58}
\usepackage{colortbl}
\usepackage{array}

\usepackage[most]{tcolorbox}
\definecolor{BoxBackground}{RGB}{240, 240, 240} 
\definecolor{BoxFrame}{RGB}{0, 0, 0} 
\definecolor{TitleBackground}{RGB}{0, 0, 0} 
\definecolor{TitleText}{RGB}{255, 255, 255} 
\tcbset{
  academicbox/.style={
    boxsep=5pt,
    left=2pt,
    right=2pt,
    bottom=0.5pt,
    boxrule=0.5pt,
    colback=BoxBackground,
    colframe=BoxFrame,
    colbacktitle=TitleBackground,
    coltitle=TitleText,
    enhanced,
    attach boxed title to top left={yshift=-0.1in,xshift=0.1in},
    boxed title style={boxrule=0pt,colframe=white},
    title={#1},
  }
}
\newtcolorbox{AcademicBox}[1][]{academicbox=#1}

\title{Tstars-Tryon 1.0: \\Robust and Realistic Virtual Try-On for Diverse Fashion Items}

\author{
{\small \textit{Pailitao Team: authors are listed in alphabetical order}} \\ \vspace{1mm}
Mengting Chen$^{*}$, Zhengrui Chen, Yongchao Du, Zuan Gao, Taihang Hu, Jinsong Lan, Chao Lin, Yefeng Shen, Xingjian Wang, Zhao Wang, Zhengtao Wu, Xiaoli Xu, Zhengze Xu, Hao Yan, Mingzhou Zhang, Jun Zheng, Qinye Zhou, Xiaoyong Zhu, Bo Zheng$^{\dagger}$ \\ \vspace{2mm}
\textbf{Alibaba Group} \\ \vspace{2mm}
{\small $^{*}$Project Lead \qquad $^{\dagger}$Corresponding Author}
}

\begin{document}

\maketitle

\begin{abstract}

Recent advances in image generation and editing have opened up new opportunities for virtual try-on applications. 
However, existing methods still struggle to meet the diverse and complex user demands in real-world scenarios. 
In this work, we present \ours, a commercial-scale virtual try-on system that is robust, realistic, versatile, and highly efficient. 
First, our system maintains a high success rate across challenging real-world cases, including extreme poses, severe illumination variation, motion blur, and other hard in-the-wild conditions. 
Second, it delivers highly photorealistic results with rich fine-grained details, faithfully preserving garment texture, material properties, and structural characteristics, while largely avoiding the synthetic artifacts often seen in AI-generated images. 
Third, beyond apparel try-on, our model serves as a general-purpose framework that supports flexible multi-image composition (up to 6 reference images) across 8 categories of fashion items: tops, pants, skirts, dresses, coats, shoes, bags, and hats, together with coordinated control over person identity and background content. 
Fourth, to overcome the latency bottlenecks of commercial deployment, our system is heavily optimized for inference speed, delivering the near real-time generation required for a seamless and interactive user experience. 
These capabilities are enabled by an integrated system design spanning end-to-end model architecture, a scalable data engine, robust infrastructure, and a carefully engineered multi-stage training paradigm. 
Extensive evaluation and large-scale product deployment demonstrate that \ours achieves leading overall performance. 
To support future research and development, we also plan to release a comprehensive benchmark. 
The model has been further extended and deployed at an industrial scale on the Taobao App, serving millions of users with tens of millions of try-on requests. 
It effectively addresses the long-standing cost–quality trade-off in e-commerce virtual try-on.
\end{abstract}

\begin{figure}[H]
    \centering
    \includegraphics[width=0.9\linewidth]{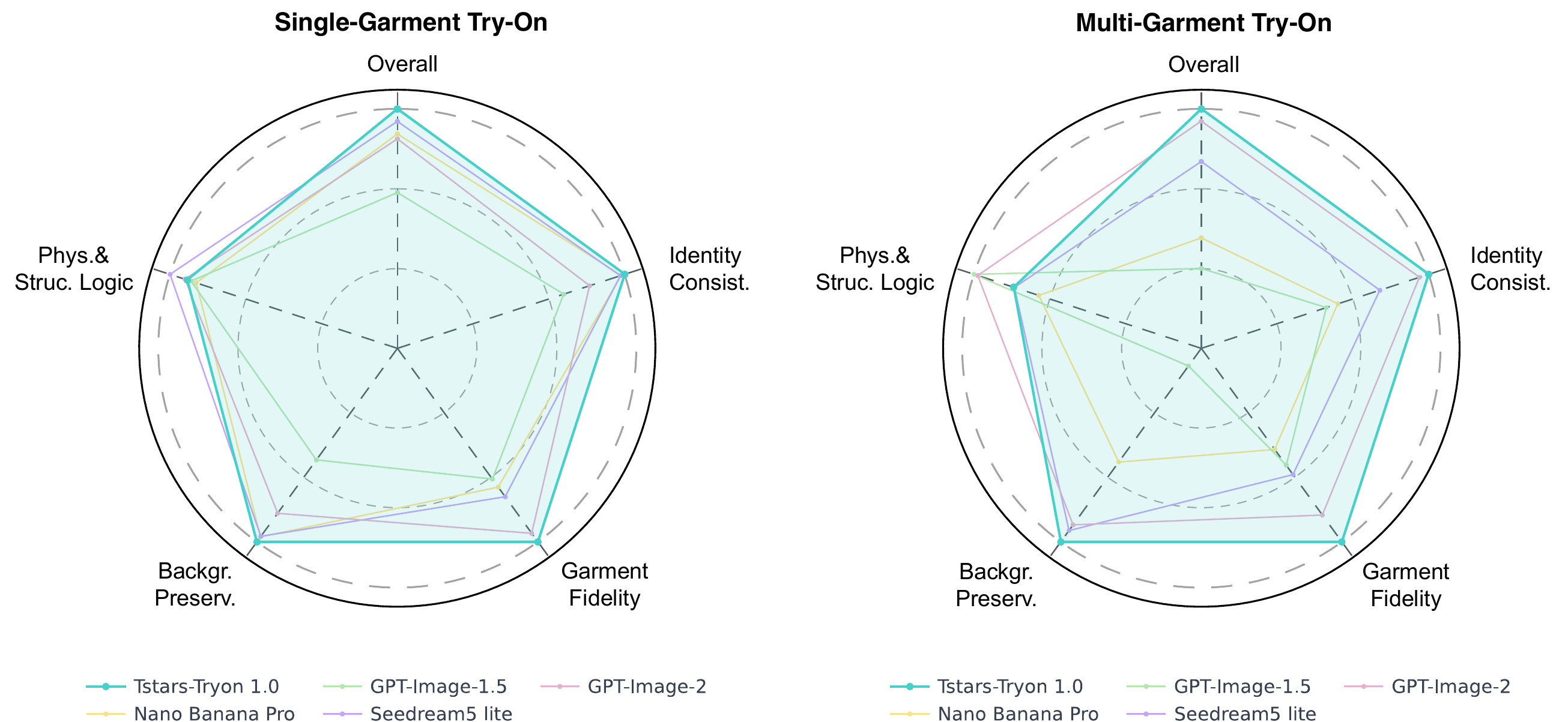}
    \caption{\textbf{Overall comparisons with state-of-the-art models.} 
    }
    \label{fig:metric}
\end{figure}

\newpage

\input{Sections/introduction}

\input{Sections/evaluation}

\input{Sections/demonstrations}

\input{Sections/deployment}

\input{Sections/acknowledgments}

\clearpage
\bibliography{colm2024_conference}
\bibliographystyle{colm2024_conference}

\end{document}

%% file: Sections/introduction.tex
\begin{figure}[H]
  \centering
  \makebox[\linewidth][c]{%
    \includegraphics[width=1.05\linewidth]{figures/teaser8-2_compressed.pdf}%
  }
  \caption{\textbf{\ours supports robust and realistic virtual try-on in the wild.} }
  \label{fig:teaser1}
\end{figure}

\begin{figure}[H]
  \centering
  \makebox[\linewidth][c]{%
    \includegraphics[width=1.05\linewidth]{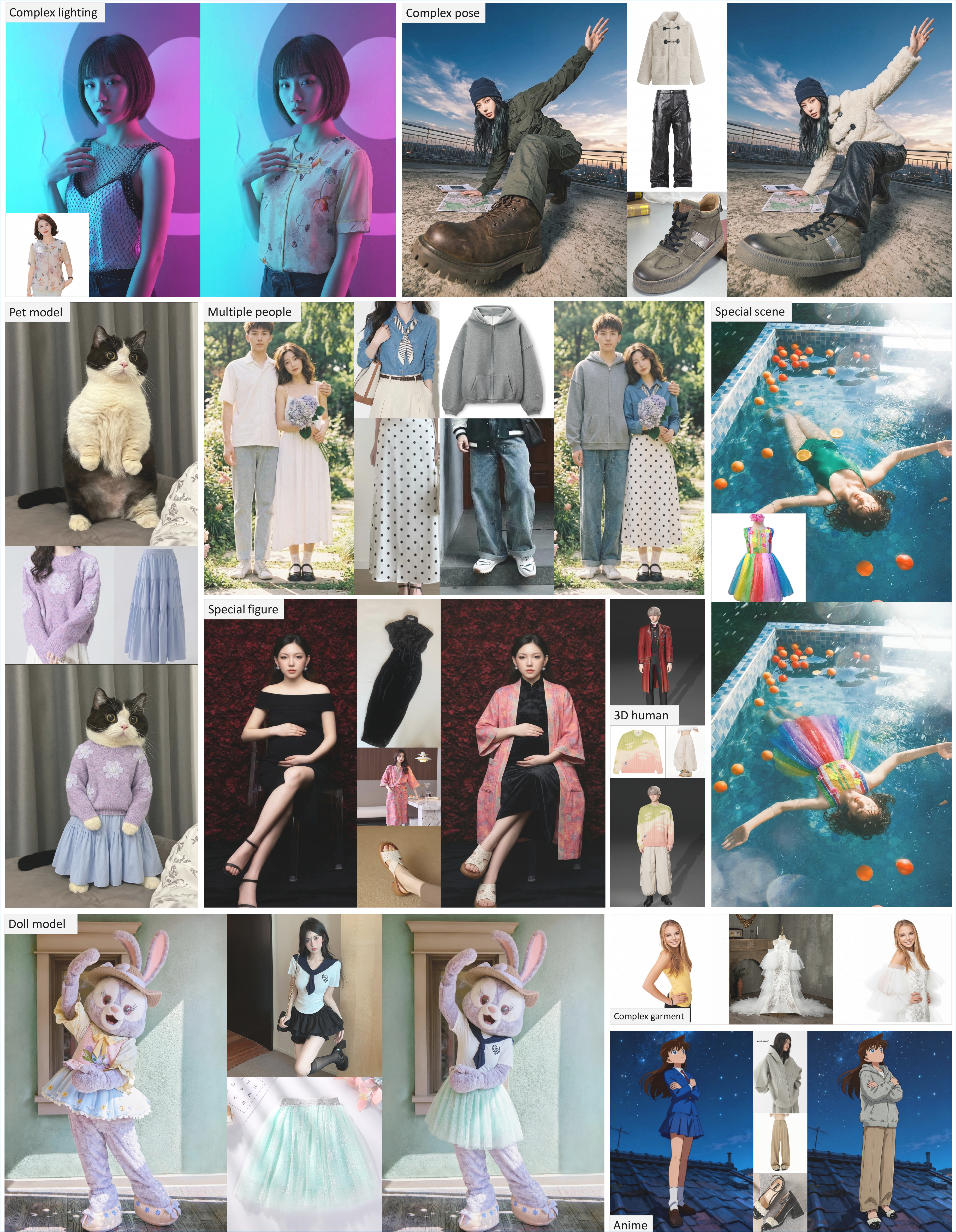}%
  }
  \caption{\textbf{\ours supports multiple challenging extreme and complex scenarios virtual try-on.}}
  \label{fig:teaser2}
\end{figure}

\section{Introduction}


Virtual try-on has emerged as one of the most compelling applications of generative AI, promising to reshape the e-commerce experience by allowing users to visualize garments on themselves before purchase. An ideal system should handle arbitrary user photos, faithfully preserve garment details, support flexible multi-item styling, and deliver results in near real-time, setting a high bar that demands continued advancement in generative modeling.

Recent advances in diffusion-based generation~\citep{rombach2022high, ho2020denoising, esser2024scaling} have substantially accelerated progress toward an ideal virtual try-on system. 
On the one hand, powerful general-purpose image editing models, both proprietary ones~(e.g., Nano Banana Pro~\citep{google2025nanobanana}, GPT-Image-1.5~\citep{gptimage15_model_card}, Seedream 5.0 lite~\citep{bytedance2026seedream}), GPT-Image-2~\citep{gptimage2_model_card},  
and open-sourced solutions~(e.g., QwenEdit-2511~\citep{wu2025qwen}, Flux-kelin~\citep{blackforest2025flux2klein}, FireRed-Image-Edit~\citep{firered2026rededit, cao2025hunyuanimage}, exhibits remarkable capabilities in complex semantic understanding and high-fidelity manipulation.
These models can directly support virtual try-on tasks or serve as robust foundation models for task-specific fine-tuning.
On the other hand, academia keeps exploring task-specific try-on models from various perspectives, such as spatial alignment and identity preservation~\citep{chong2024catvtonconcatenationneedvirtual, jiang2024fitdit, xu2025ootdiffusion}. 
Collectively, these synergistic explorations provide vital opportunities and a strong foundation for the commercialization of virtual try-on technologies.

However, moving toward true commercial-grade applications remains challenging. 
First, commercial systems demand rigorous robustness to seamlessly process diverse, in-the-wild user photos, which frequently feature extreme poses, overexposure, unconventional angles, and complex background scenes. Second, unprecedented realism is crucial, necessitating the extreme restoration and exact preservation of intricate garment details and fabrics. Third, true flexibility requires moving beyond single-item scenarios to seamlessly support multi-image inputs, cross-category generation, and complex layering or outfit combinations. Finally, real-world deployment imposes stringent demands on inference speed, requiring near real-time generation to provide instant feedback and ensure a seamless user experience. Given these demanding criteria, existing methods still exhibit a notable gap toward realizing true commercial-grade applications.

\begin{figure}[h]
    \centering
    \includegraphics[width=\linewidth]{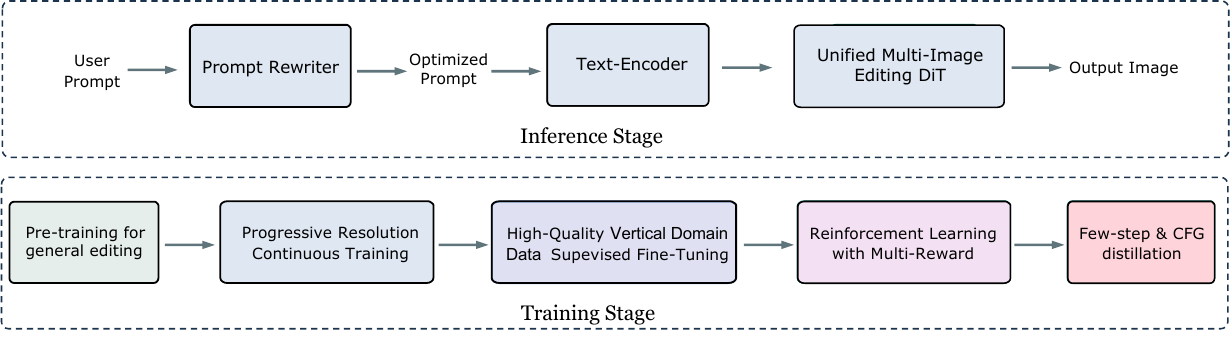}
    \caption{\textbf{Overview of training and inference pipeline of \ours.}}
    \label{fig:pipeline}
\end{figure}
To overcome these limitations, we reformulate the full-stack pipeline for a commercial-level foundation model, from data curation, model architecture, to training strategies and inference optimization. The overall framework is shown in Figure~\ref{fig:pipeline}:
%

\begin{figure}[t]
    \centering
    \includegraphics[width=\linewidth]{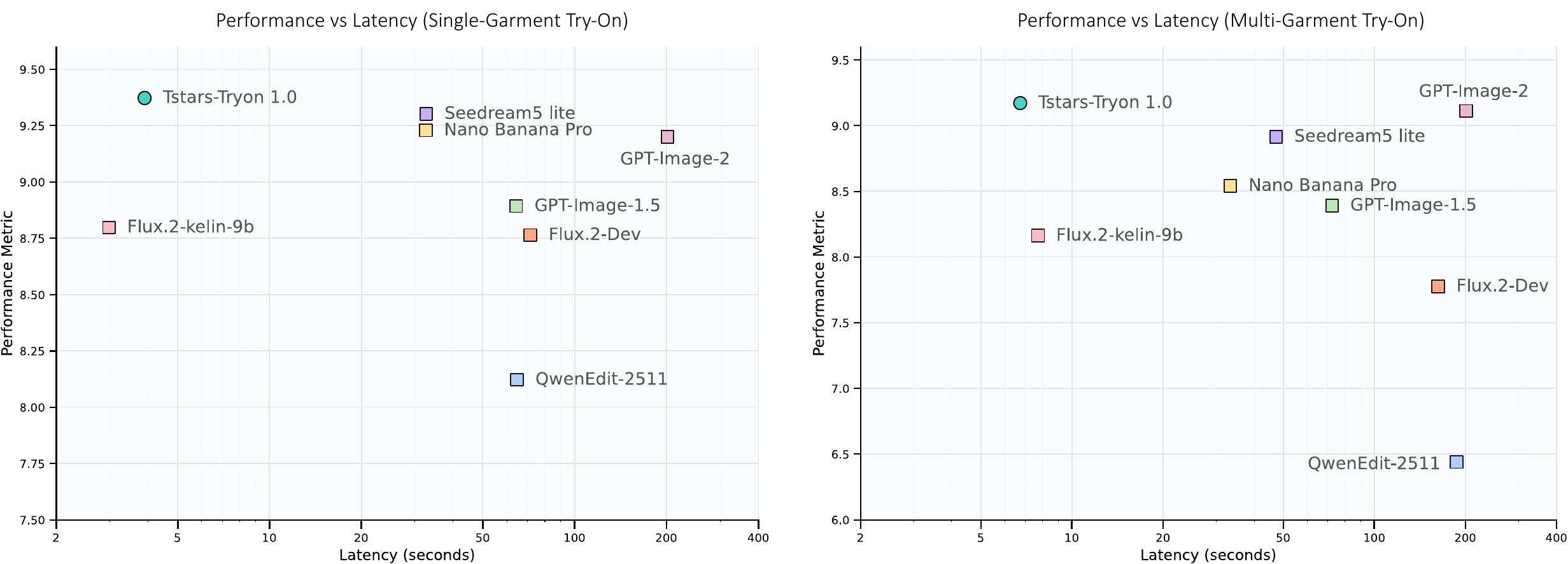}
    \caption{\textbf{Performance and latency evaluation on the Tstars-VTON Benchmark.} \ours achieves optimal performance in the single-garment scenario with a rapid 3.92s latency. For complex multi-garment try-on (5 reference images in average), it still delivers outputs in just 6.74s. Meanwhile, top open-source models (QwenEdit-2511\citep{wu2025qwen}, Flux.2 dev\citep{blackforest2025flux2klein}) take $\sim$200s. Note: Tested on an H200 GPU. Closed-source model times are estimated via API calls and may include network latency (for reference only).}
    \vspace{-4mm}
    \label{fig:latency}
\end{figure}

\begin{itemize}[leftmargin=*]
    \item 
    \textbf{Data Engine}: To address the scarcity of multi-item try-on data, we built an automated pipeline for large-scale, high-quality image editing datasets. This workflow integrates image element decomposition and retrieval-based recall systems to build a robust data pool. We utilize customized captioners for professional-grade descriptions, further refined by knowledge-enhanced Vision Language Model post-filtering and extensive perceptual metric screening.
    \item 
    \textbf{Model Architecture}: Moving away from traditional inpainting logic, we treat virtual try-on as a specialized image editing task. \ours utilizes a unified MMDiT~\citep{esser2024scaling} architecture capable of simultaneously processing and coordinating multiple reference images, ensuring the natural fusion of full-body outfits.
    \item 
    \textbf{Training Infra}: Our framework natively supports variable resolutions and an arbitrary number of reference images. By leveraging Data Parallelism, Tensor Parallelism, and adapting Data Packing strategies~\citep{dehghani2023patch} for Diffusion Transformers, we have eliminated the computational waste typically associated with traditional bucketing strategies.
    \item 
    \textbf{Meticulous Training Strategies}: During pre-training, we utilize task-balanced and content-balanced datasets with a progressive difficulty scaling strategy to bolster the model’s world knowledge and general editing capabilities. We further apply progressive resolution continuous training to enhance high-resolution synthesis. In the high-quality SFT stage, we curate and balance vertical domain data, while guiding the training process through comprehensive metric monitoring. During reinforcement learning, we perform group-level trajectory sampling and use a multi-dimensional reward pipeline to estimate each sample’s group-relative advantage. Built upon the SFT checkpoint, the policy is further optimized with DiffusionNFT~\citep{zheng2025diffusionnft} to favor positive trajectories over negative ones. This stage yields strong CFG-free inference performance and further improves garment consistency, outfit quality, and generation stability, even under complex human poses or intricate garment designs.
    \item 
    \textbf{Prompt Enhancement}: We introduced a tailored rewriter model to enhance semantic features. This model accurately identifies and describes complex virtual try-on editing processes, providing precise semantic guidance that enhances the detail and accuracy of the final output.
    \item 
    \textbf{Fast Inference Acceleration}: To meet the low-latency demands of live business environments, \ours primary DiT model is streamlined to \textbf{5B parameters}. 
    By combining CFG (Classifier-Free Guidance) distillation and Step Distillation~\citep{yin2024one}, we have achieved just \textbf{3.92 seconds} for single-garment and \textbf{6.74 seconds} for multi-garment try-on (5 reference images in average) without compromising visual fidelity. See Figure~\ref{fig:latency} for details.
    \item 
    \textbf{Business-Centric Protocol}: We developed Tstars-VTON Benchmark, a comprehensive evaluation suite to validate commercial value. This framework covers a vast array of model body types and all product categories, simulating real-world performance across a global user base and inventory.
    \end{itemize}
Relying on the underlying technical breakthroughs, this model has demonstrated powerful application scalability in actual business scenarios:
\begin{itemize}[leftmargin=*]
    \item 
    \textbf{Extreme Robustness}: Breaks through the limitations of body type and style, offering flawless support for diverse human poses, extreme lighting conditions, and any combination of complex garments in the wild.
    \item 
    \textbf{High-Fidelity Realism}: Deeply restores the unique ID and intricate material textures of clothing, ensuring that the final rendering is physically accurate, free of synthetic artifacts, and naturally "wearable."
    \item 
    \textbf{Unprecedented Flexibility}: Features industry-leading multi-item generation across eight categories (tops, pants, skirts, dresses, shoes, bags, hats, and coats). Furthermore, it transcends physical boundaries by supporting non-photorealistic inputs (e.g., digital humans or anime characters) and seamlessly integrates with complex general image editing tasks.
    \item 
    \textbf{High-Efficiency Inference}: Designed for high-concurrency online scenarios, \ours optimized architecture significantly reduces VRAM usage and latency while maintaining lossless image quality, delivering the near real-time generation required for seamless commercial deployment.
\end{itemize}
Extensive qualitative and quantitative experiments demonstrate the superiority of \ours. As illustrated by the quantitative comparisons in Figure~\ref{fig:metric} and Figure~\ref{fig:latency}, our model consistently outperforms top-tier commercial models(Nano Banana Pro\citep{google2025nanobanana}, GPT-Image-1.5\cite{gptimage15_model_card} and even the latest released model GPT-Image-2\citep{gptimage2_model_card}) across key dimensions in both single- and multi-garment try-on tasks, proving that state-of-the-art results can be achieved with significantly reduced computational overhead. Figure~\ref{fig:teaser1} showcases the model's robust capabilities for realistic virtual try-on in the wild, maintaining high visual fidelity across diverse dynamic environments and seamlessly adapting to non-standard subjects such as statues and anime characters. Furthermore, Figure~\ref{fig:teaser2} highlights the exceptional versatility of Tstars-Tryon 1.0 in a variety of complex scenarios, proving its ability to handle complex lighting, extreme poses, multiple people, special figures (such as maternity wear), and even cross-domain subjects like pets and dolls. In summary, these results demonstrate that Tstars-Tryon 1.0 establishes a new industry-leading standard for highly adaptable and high-fidelity virtual try-on generation.

%% file: Sections/evaluation.tex
\section{Tstars-VTON Benchmark}

To evaluate the model under commercial standards, we introduce \textbf{Tstars-VTON Benchmark}. This benchmark explicitly incorporates the challenges from real applications for rigorous evaluation, such as multi-garment layering, complex background, and diverse human poses.
Specifically, we collect large-scale data and refine them into 1780 paired samples across 5 garment categories and 3 accessory categories, covering 465 fine-grained subcategories and 1-6 layered try-on items.

\begin{figure}[t]
\centering
\includegraphics[width=0.99\linewidth]{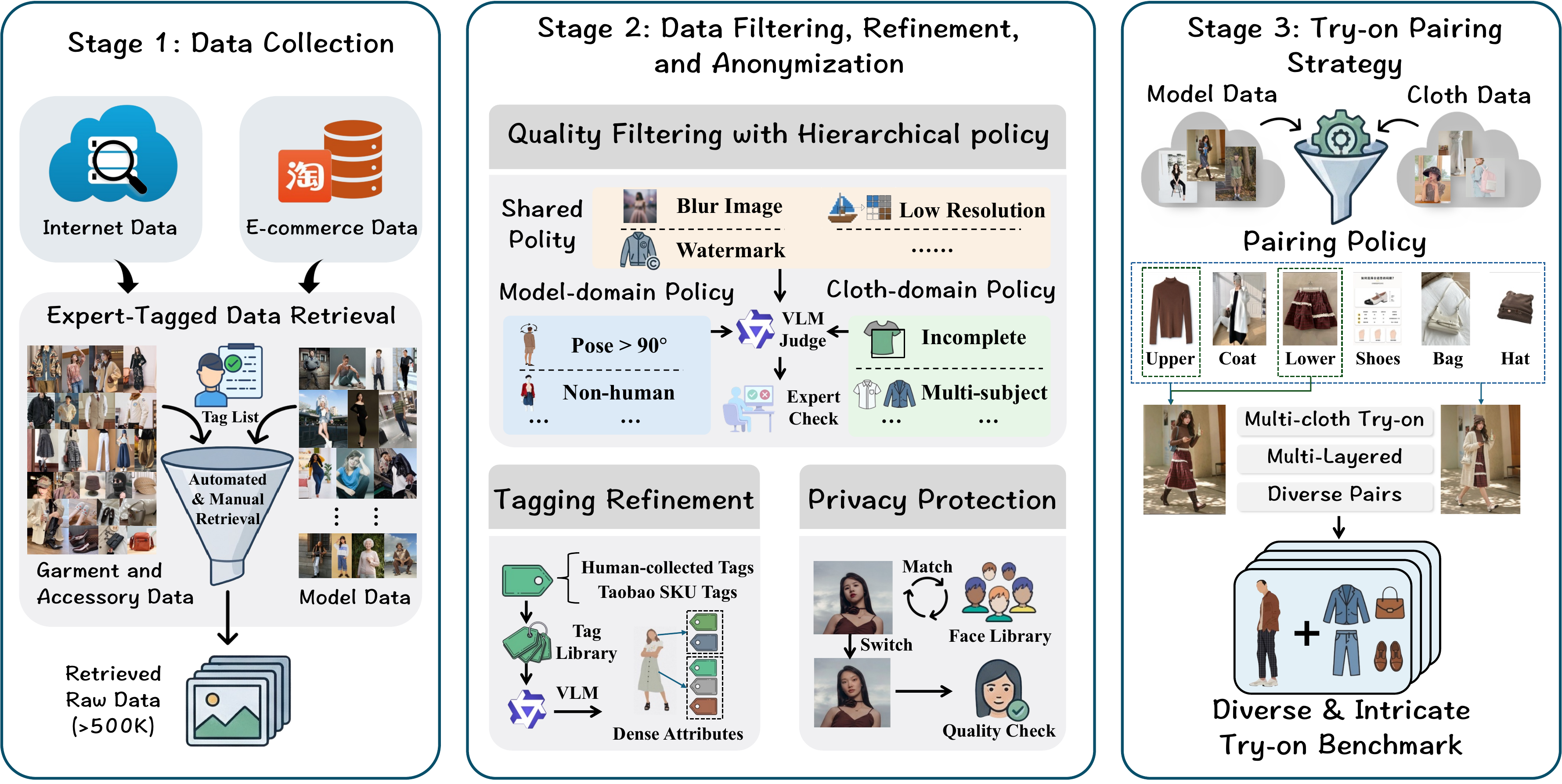}
\caption{\textbf{Data curation pipeline of Tstars-VTON Benchmark.}\label{fig:benchmark_data_curation}}
    \vspace{-4mm}
\end{figure}


\subsection{Limitations of Academic Benchmarks}

Despite their widespread adoption, existing academic benchmarks exhibit significant limitations that hinder the evaluation of models for real-world deployment. First, they suffer from homogeneous backgrounds and restricted garment categories. Datasets such as VITON-HD~\citep{choi2021vitonhd} and DressCode~\citep{morelli2022dresscode} predominantly feature simplistic studio-like backgrounds and confine their scope to basic topological categories, such as upper, lower, and dresses. In contrast, authentic e-commerce and daily-life scenarios encompass complex in-the-wild environments. Furthermore, a comprehensive virtual try-on experience frequently involves a broader array of fashion items, including outerwear, footwear, bags, and accessories (e.g., hats), which are largely overlooked by current benchmarks.

Second, existing datasets fail to reflect the diversity and complexity of reference garment images provided by users in practical applications. Most academic benchmarks are strictly designed for single-garment try-on. While recent efforts, such as DressCode-MR~\citep{chong2025fastfitacceleratingmultireferencevirtual}, attempt to address multi-garment scenarios, the reference garment images in these datasets are often artificially extracted from the source model images. More critically, existing benchmarks implicitly assume that reference garments are pristine flat-lay images on simple backgrounds. However, user-provided reference images are highly unconstrained in real-world commercial scenarios. The user-provided reference images frequently feature complex backgrounds or are even in-the-wild portrait photos of other individuals.

Consequently, evaluating models on these constrained datasets fails to reflect their true capability in handling the intricate garment combinations and complex reference conditions required for industrial deployment. To comprehensively assess whether a virtual try-on model is truly capable of functioning in real-world scenarios, a new highly practical benchmark is urgently needed.

\subsection{Principles of Our Benchmark}


To address the limitations above, we construct a new benchmark that truly meets the practical demands with the following distinctive features.

\begin{figure}[t]
\centering
\includegraphics[width=\linewidth]{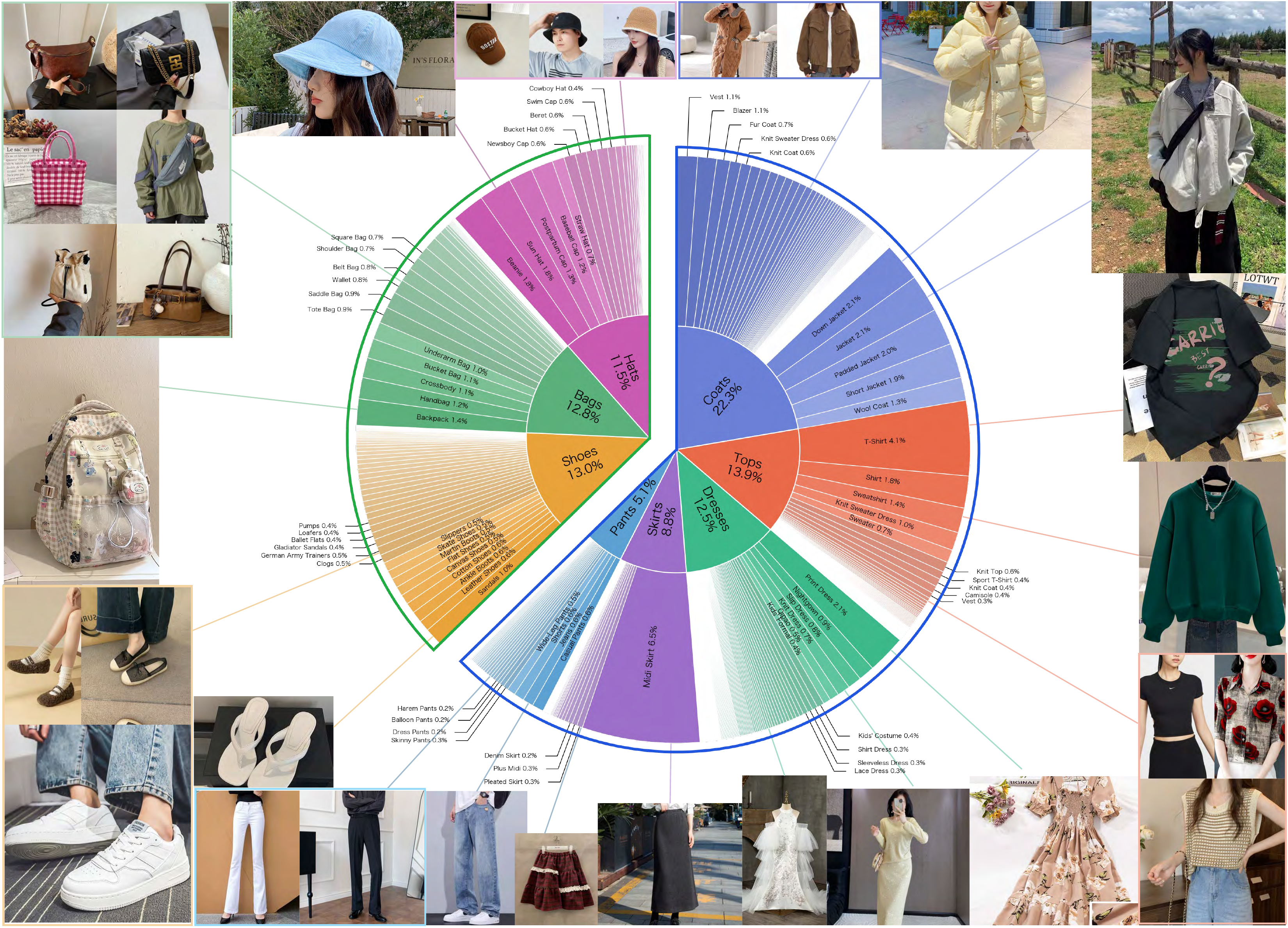}
\caption{\textbf{Clothing statistics of Tstars-VTON Benchmark.} The distributions of garments and accessories are illustrated in two separate fan-shaped arrangements with blue and green borders, respectively, while representative images are shown around.\label{fig:benchmark_cloth_stats}}
    \vspace{-4mm}

\end{figure}

\begin{itemize}[leftmargin=*]
    \item \textbf{Multi-Garment/Accessory Try-On Scenarios}: Unlike most prior benchmarks that focus on single-garment try on~\citep{choi2021vitonhd,morelli2022dresscode}, the Tstars-VTON benchmark introduces multi-garment combinations with layered outfits and accessories to capture real-world dressing complexity, covering free-combination scenarios with 1-6 items.
    This substantially increases scenario diversity and enables a more thorough evaluation of generative models in multi-condition controllable synthesis and semantic understanding.
    \item \textbf{Diverse Data Coverage with Fine-Grained Attributes}: 
    To overcome data coverage limitation caused by source distribution bias~\citep{choi2021vitonhd,morelli2022dresscode}, our benchmark collects diverse data from Internet and E-commerce domains, following by a two-stage tag retrieval and refinement process involving VLM-based generation and manual check.
    Our benchmark adopts controllable data sampling over multiple tag attributes, including 11 tag dimensions for models and 13 tag dimensions for garments, to realize broad and balanced data coverage.
    \item \textbf{Privacy-Preserving Mechanism}: 
    All portraits collected from open-source data are matched to the most similar models in a licensed face database and anonymized through face swapping, following by a rigorous verification to ensure swapping quality.
    \item \textbf{Flexible Unpaired Settings}: Tstars-VTON benchmark supports a fully unpaired evaluation setting, decoupling the model and garment databases to maximize combinatorial diversity while maintaining control over attribute pairing process.
    \item \textbf{Comprehensive Evaluation Paradigm Aligned with Human Preferences}: 
    We propose a VLM-driven evaluation paradigm that decomposes virtual try-on quality into four rigorous dimensions, each evaluated on a 1-10 Likert scale. 
    This framework supports both single-garment and multi-garment scenarios by assessing Identity Consistency, Garment Fidelity, Background Preservation, Physical and Structural Logic. 
    By integrating human-aligned prompts and interpretable scoring rationales, this protocol bridges automated metrics with fine-grained diagnostic assessment, ensuring the evaluation paradigm meets commercial-grade standards.
\end{itemize}

\begin{figure}[t]
    \centering
    \begin{subfigure}[b]{0.3\textwidth}
        \centering
        \includegraphics[width=\textwidth]{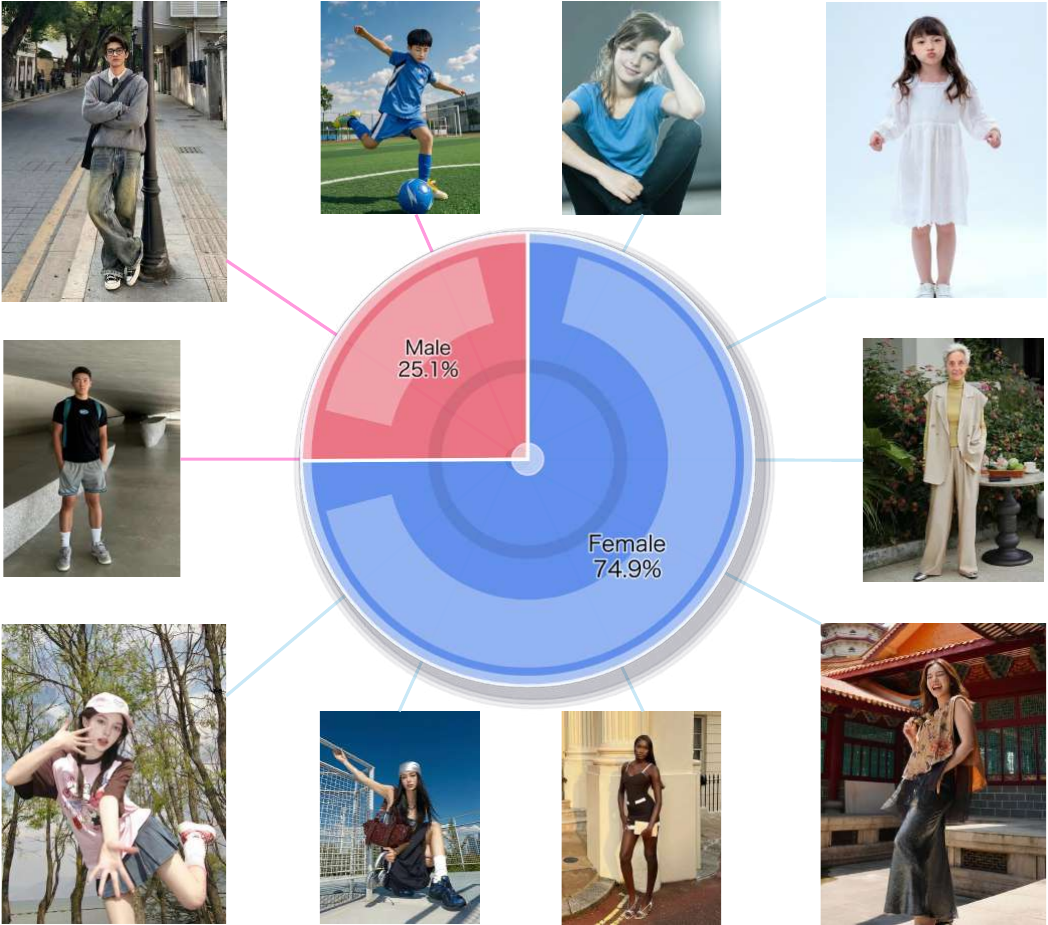}
        \caption{Distribution of model genders.}
    \end{subfigure}
    \hfill
    \begin{subfigure}[b]{0.3\textwidth}
        \centering
        \includegraphics[width=\textwidth]{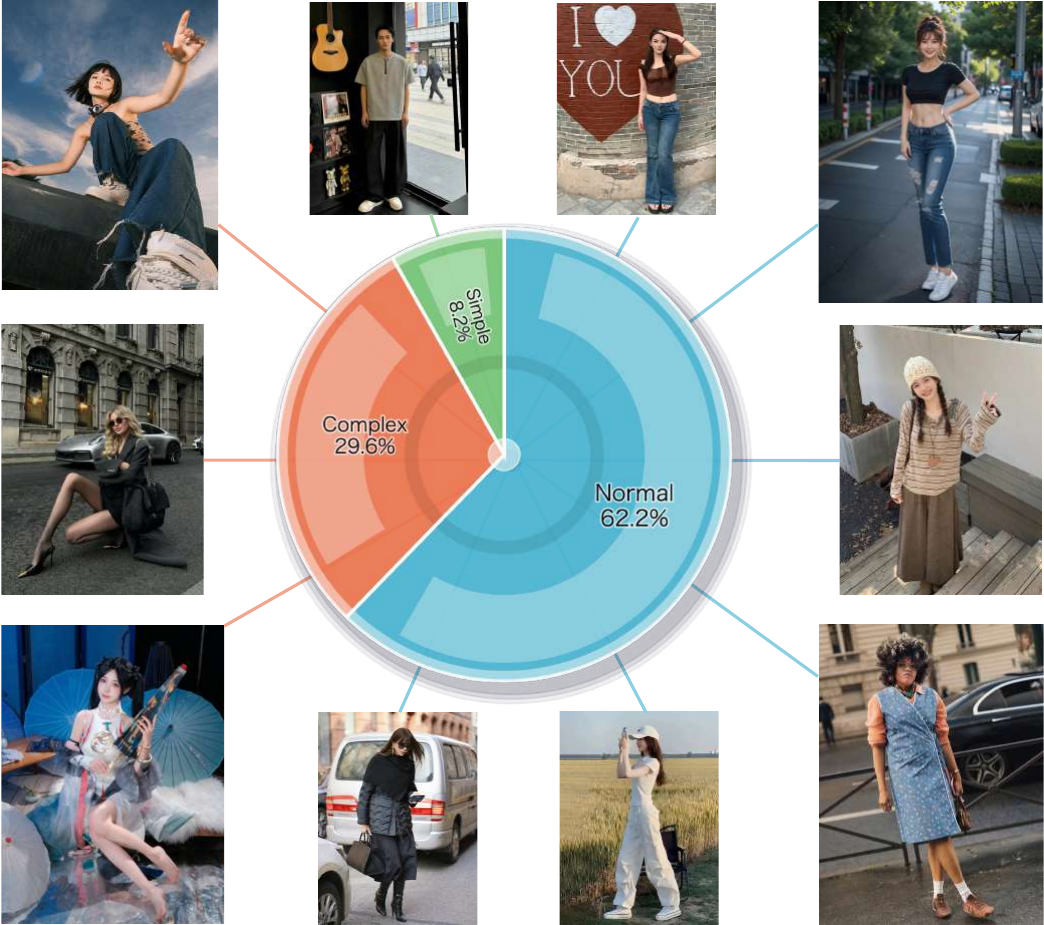}
        \caption{Distribution of model poses.}
    \end{subfigure}
    \hfill
    \begin{subfigure}[b]{0.3\textwidth}
        \centering
        \includegraphics[width=\textwidth]{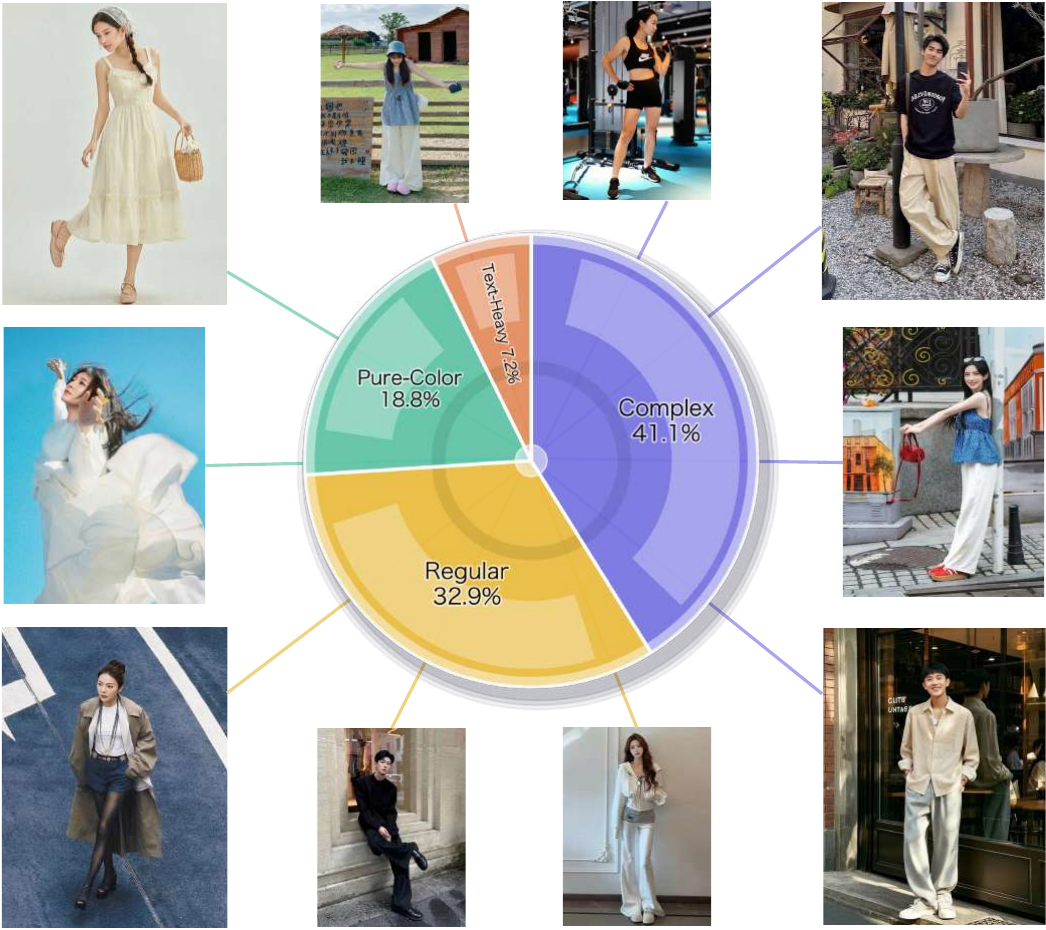}
        \caption{Distribution of scenarios.}
    \end{subfigure}
    \caption{\textbf{Model statistics of Tstars-VTON benchmark.} Pose diversity and scenario variety along with basic model attributes are captured. Each subfigure is supplemented with representative images corresponding to specific attributes.\label{fig:benchmark_model_stats}}
        \vspace{-4mm}

\end{figure}

\subsection{Benchmark Construction}
Following above principles, we construct the Tstars-VTON benchmark through a three-stage pipeline to optimize the quality and diversity of paired samples under realistic try-on constraints, as shown in Figure \ref{fig:benchmark_data_curation}.

\noindent\textbf{Stage 1. Data Collection.}
To achieve broad data coverage and reduce the distribution biases common in existing benchmarks, we build a comprehensive data pool consisting of two image sets, i.e., diverse human models and a wide range of garments and accessories.
We retrieve large-scale data from internet and proprietary e-commerce sources.
To better reflect real-world try-on distributions, human experts first design a multi-dimensional tag system.
Guided by these tags, we adopt a hybrid retrieval strategy that combines automated platform extraction with targeted manual collection, covering diverse model characteristics, dressing poses, and complex garment/accessory types.

\noindent\textbf{Stage 2. Data Filtering, Refinement, and Anonymization.}
After collection, the raw data are rigorously filtered by fine-grained rules and refined to ensure quality.
This process combines automated tools, such as a VLM-based pipeline, with human verification.
Then each image is assigned rich semantic tags through a multi-step annotation process.
Labels are first derived from internet or e-commerce SKU metadata and manually verified.
A VLM further refines the annotations and supplements fine-grained attributes, followed by final manual checking to ensure consistency and accuracy.

For privacy protection, all model portraits undergo face swapping.
Each face is matched to a licensed surrogate by attributes such as skin tone, gender, and age.
Reference-guided swapping improves realism, while failed cases are iteratively corrected through automated filtering and human inspection.

\noindent\textbf{Stage 3. Try-on Pairing Strategy.}
The pairing process is aiming to maximize the diversity of matching selections, while strictly adhering to realistic physical and semantic rules.
Beyond simple constraints such as gender-matching, the outfit combinations follow a structured layering logic, which enforces unique image utilization and establishes precise coexistence protocols.
This strategy dynamically generates diverse, physically plausible layered outfits that accurately mirror real-world dressing complexity.

\subsection{Evaluation Metrics}
To move beyond simplistic automated metrics (e.g., FID~\citep{heusel2017gans}) that often fail to capture fine-grained visual defects, we introduce a specialized VLM-driven evaluation protocol. This protocol decomposes try-on quality into four semantically distinct dimensions, each scored on a scale of 1 to 10. To maximize evaluative precision and prevent the model from being distracted by irrelevant information, the protocol is executed through two independent API calls with different image inputs:

\begin{itemize}[leftmargin=*]
\item \textbf{Stage 1: Garment-Aware Evaluation}. This stage provides the VLM with the original person, the reference garment(s), and the result. It focuses on the relationship between the target items and the subject.
\begin{itemize}
\item \textit{Identity Consistency}: This dimension evaluates the preservation of the person's face, pose, and body shape. It is intentionally included in Stage 1 because the reference garments provide critical context for the person's silhouette. By seeing the clothing (e.g., a bulky jacket versus a tight top), the VLM can correctly distinguish whether a change in the person's visible scale or shape is a natural result of the outfit's style or an actual failure to preserve the subject's identity.
\item \textit{Garment Fidelity}: For both single and multi-garment tasks, the VLM evaluates each target item individually based on its category label, ensuring that the silhouette, texture, and complex patterns of the reference are faithfully reproduced without being influenced by other parts of the outfit.
\end{itemize}
\item \textbf{Stage 2: Garment-Agnostic Evaluation}. In this stage, the VLM only receives the original person and the result image, without the garment references. This isolation allows the model to focus exclusively on background integrity and structural realism.
\begin{itemize}
    \item \textit{Background Preservation}: The VLM first performs a background type classification (Plain vs. Complex). The score is then based on color consistency (for plain backgrounds) or pixel-level content preservation and lighting consistency (for complex backgrounds).
    \item \textit{Physical and Structural Logic}: This dimension monitors anatomical correctness and mesh clipping. To minimize false positives in complex poses, the VLM must perform a second verification pass before flagging limb anomalies. It also checks for interpenetration, such as fabric visibly passing through skin or layers of clothing intersecting in physically impossible ways.
\end{itemize}
\end{itemize}

\paragraph{Overall Score Calculation.}
To synthesize these four dimensions into a single representative metric, we compute the \textbf{Overall Score} using the \textit{geometric mean} of the individual scores. Unlike the standard arithmetic mean, the geometric mean is highly sensitive to "weak links." This ensures that a model must achieve balanced excellence across all categories.

\subsection{Benchmark Statistics}
Tstars-VTON benchmark provides diverse try-on pairs whose distributions are consistent with real-world scenarios.
Detailed statistics are introduced as follows.

\begin{figure}[t]
    \centering
    \begin{subfigure}[b]{0.24\textwidth}
        \centering
        \includegraphics[width=\textwidth]{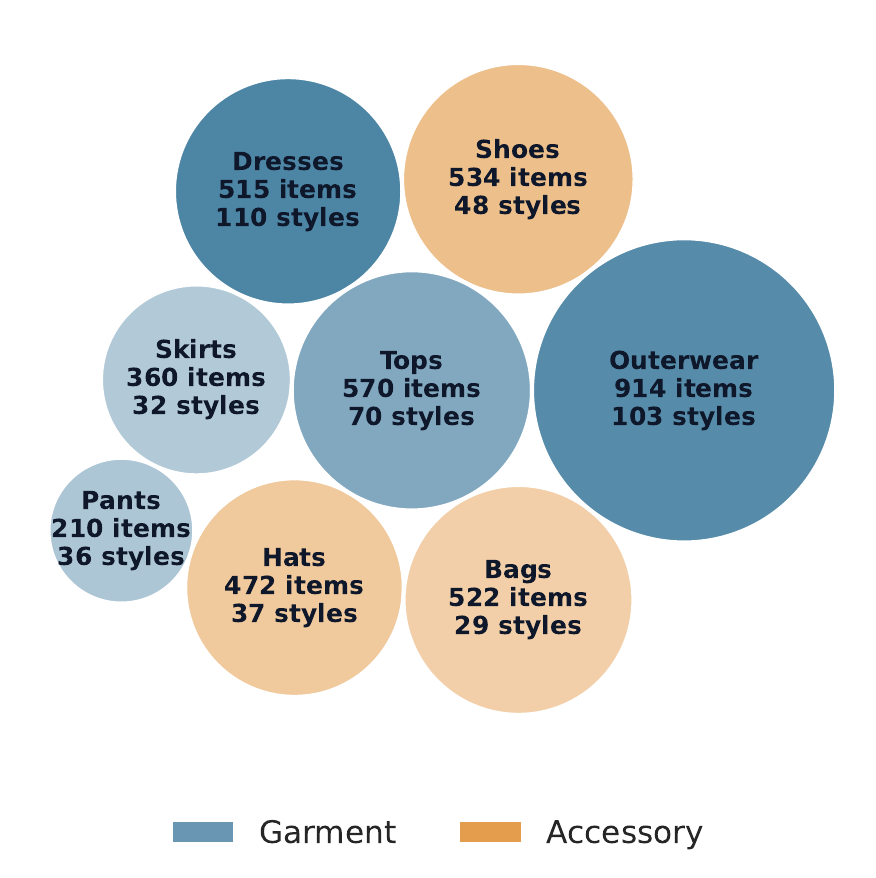}
        \caption{Distribution of garment/accessory sub-categories.}
    \end{subfigure}
    \hfill
    \begin{subfigure}[b]{0.72\textwidth}
        \centering
        \includegraphics[width=\textwidth]{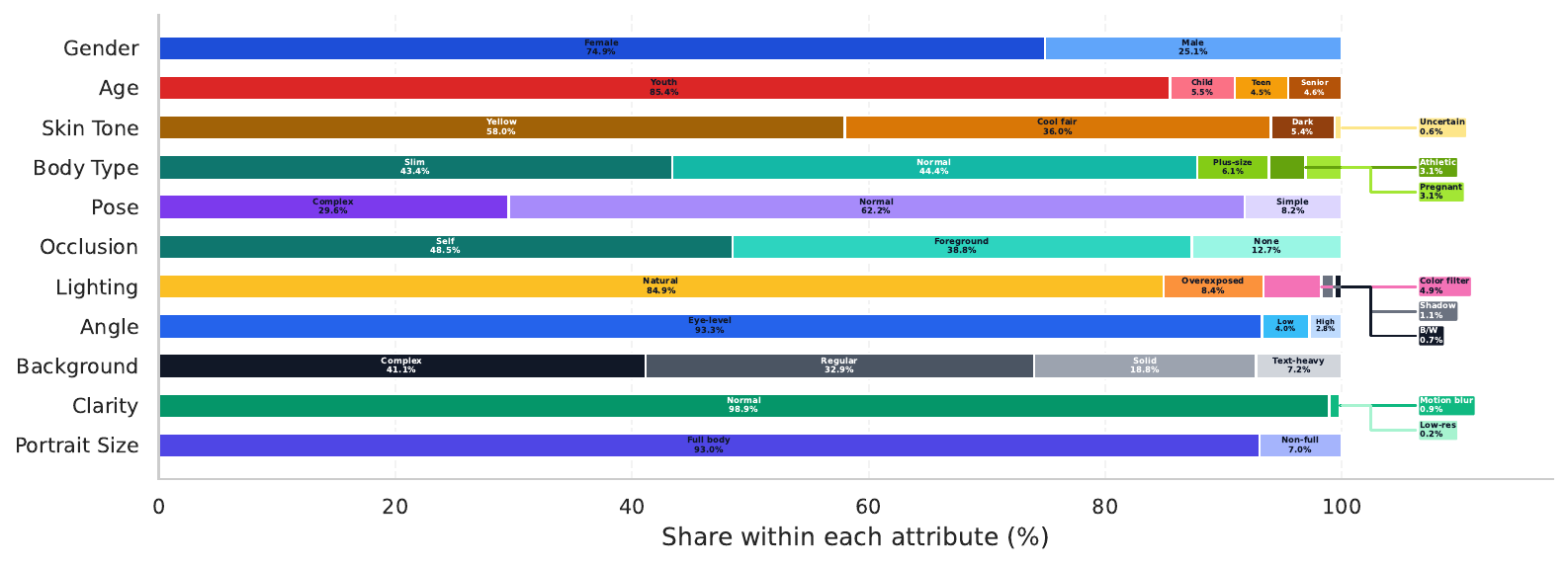}
        \caption{Distribution of model characteristics.}
    \end{subfigure}
    \caption{\textbf{Attribute statistics of Tstars-VTON benchmark.} Diversity of clothing and model attributes is shown in sub-figures above.\label{fig:benchmark_detail_stats}}
\end{figure}

\noindent\textbf{Distribution of Garments and Accessories.}
To closely mirror realistic try-on complexity, as shown in Figure \ref{fig:benchmark_cloth_stats} and left part of Figure \ref{fig:benchmark_detail_stats}, the dataset incorporates 5 garment categories and 3 accessory categories, which are further subdivided into totally 465 fine-grained sub-styles.
For garments, there are tops with 70 styles, dresses with 110 styles, coats with 103 styles, pants with 36 styles, and skirts with 32 styles.
And for accessories, there are shoes with 48 styles, hats with 37 styles, and bags with 29 styles.
This diverse collection supports complex multi-item try-on tasks ranging from 1 to 6 items per sample.

\noindent\textbf{Distribution of Model Characteristics.}
We carefully curated the dataset to incorporate diverse human characteristics, varied poses, and in-the-wild backgrounds, as shown in Figure \ref{fig:benchmark_model_stats} and right part of Figure \ref{fig:benchmark_detail_stats}.
In terms of demographic diversity, the dataset includes models across different genders including 74.9\% female and 25.1\% male, and varied age groups featuring 85\% youth but also encompassing children, teenagers, and seniors.
To further challenge the try-on models, the dataset incorporates 29.6\% proportion of complex poses rather than focusing solely on simple poses which only accounts for 8.2\%.
Furthermore, more than 40\% of model images feature highly complex backgrounds for intricate in-the-wild settings.

\section{Evaluation Results}

\subsection{Quantitative Results}
\label{sec:quant_results}

\begin{table}[t]
\centering
\caption{\textbf{Quantitative results on the Tstars-VTON Benchmark (Single-Garment).} The best and second-best results are demonstrated in \textbf{bold} and \underline{underlined}, respectively.}
\label{tab:taobao_single_results}
\resizebox{0.9\linewidth}{!}{
\begin{tabular}{lccccc}
\toprule
Method & Overall $\uparrow$ & \makecell{Identity \\ Consist.} & \makecell{Garment \\ Fidelity} & \makecell{Backgr. \\ Preserv.} & \makecell{Phys. \& \\ Struc. Logic} \\
\midrule
\rowcolor[gray]{0.95} \textit{Academic SOTA} & & & & & \\
CatVTON \citep{chong2024catvtonconcatenationneedvirtual} & 6.663 & 9.335 & 4.007 & 9.474 & 7.955 \\
Leffa \citep{zhou2024learning} & 6.048 & 8.135 & 4.348 & 8.753 & 6.009 \\
FitDiT \citep{jiang2024fitdit} & 5.152 & 6.767 & 4.706 & 8.028 & 3.883 \\
FastFit \citep{chong2025fastfitacceleratingmultireferencevirtual} & 6.448 & 9.131 & 4.672 & 8.338 & 6.546 \\
\midrule
\rowcolor[gray]{0.95} \textit{Proprietary / Gen-Edit} & & & & & \\
QwenEdit-2511 \citep{wu2025qwen} & 8.121 & 9.214 & 6.787 & 9.168 & 8.865 \\
FLUX.2-dev \citep{blackforest2025flux2klein} & 8.764 & 9.419 & 7.920 & 9.640 & 8.960 \\
FLUX.2-klein-9B \citep{blackforest2025flux2klein} & 8.797 & 9.442 & 8.183 & 9.504 & 8.902 \\
FireRed-Image-Edit-1.1 \citep{firered2026rededit} & 8.863 & 9.610 & 7.796 & 9.775 & 9.068 \\
GPT-Image-1.5$^{\dagger}$ \citep{gptimage15_model_card} & 8.892 & 9.381 & 8.563 & 9.075 & 9.219 \\
GPT-Image-2$^{\dagger}$ \citep{gptimage2_model_card}& 9.200 & 9.597 & \underline{8.794} & 9.588 & \underline{9.255} \\
Nano Banana Pro \citep{google2025nanobanana} & 9.229 & \underline{9.861} & 8.598 & \underline{9.816} & 9.189 \\
Seedream5 lite \citep{seedream2025seedream}& \underline{9.301} & 9.854 & 8.639 & 9.810 & \textbf{9.343} \\
\midrule
\textbf{\ours} & \textbf{9.372} & \textbf{9.889} & \textbf{8.833} & \textbf{9.863} & 9.241 \\
\bottomrule
\end{tabular}
}
\vspace{1mm}
\begin{minipage}{0.9\linewidth}
\scriptsize $^{\dagger}$ Due to platform restrictions, GPT-Image-1.5 failed to generate results for 120 test cases and GPT-Image-2 failed to generate results for 107 test cases. The reported metrics are calculated excluding these missing instances. Both GPT models were queried via the official API with the quality parameter set to \texttt{high}.
\end{minipage}
\end{table}

To further validate the commercial utility of our framework, we conduct a comprehensive evaluation on the Tstars-VTON Benchmark, covering both single-garment and multi-garment scenarios. Unlike traditional academic benchmarks, this dataset presents significant challenges in terms of pose diversity, background complexity, and high-fidelity texture requirements. The results are summarized in Table \ref{tab:taobao_single_results} and Table \ref{tab:taobao_multi_results}.

\paragraph{Single-Garment Try-On.}
As shown in Table \ref{tab:taobao_single_results}, there is a clear performance hierarchy between specialized academic models and general-purpose editing models. Current academic state-of-the-art (SOTA) models exhibit limited robustness in complex scenarios, often failing to produce high fidelity garments.

While leading proprietary models such as GPT-Image-1.5, Nano Banana Pro, and Seedream5 lite provide strong competition, \ours consistently achieves superior or competitive performance across all dimensions. Specifically, \ours demonstrates a clear advantage in \textit{Garment Fidelity}, effectively capturing intricate textures, material drapes, and fine patterns that competing models sometimes blur or over-simplify. Furthermore, our model excels in \textit{Identity Consistency} and \textit{Background Preservation}, ensuring that the subject's facial features and the original environment remain pixel-identical to the source, even when the new garment introduces significant changes to the person's silhouette.

\begin{table}[t]
\centering
\caption{\textbf{Quantitative results on the Tstars-VTON Benchmark (Multi-Garment).}}
\label{tab:taobao_multi_results}
\resizebox{0.9\linewidth}{!}{
\begin{tabular}{lccccc}
\toprule
Method & Overall $\uparrow$ & \makecell{Identity \\ Consist.} & \makecell{Garment \\ Fidelity} & \makecell{Backgr. \\ Preserv.} & \makecell{Phys. \& \\ Struc. Logic} \\
\midrule
\rowcolor[gray]{0.95} \textit{Open-source} & & & & & \\
FastFit \citep{chong2025fastfitacceleratingmultireferencevirtual} & 6.039 & 8.163 & 4.575 & 8.096 & 5.847 \\
QwenEdit-2511 \citep{wu2025qwen} & 6.441 & 7.274 & 5.638 & 7.256 & 8.235 \\
FLUX.2-dev \citep{blackforest2025flux2klein} & 7.775 & 7.964 & 7.797 & 8.508 & 8.458 \\
FLUX.2-klein-9B \citep{blackforest2025flux2klein} & 8.161 & 8.711 & 7.870 & 8.979 & 8.363 \\
FireRed-Image-Edit-1.1 \citep{firered2026rededit} & 4.822 & 5.393 & 4.837 & 4.879 & 5.139 \\
\midrule
\rowcolor[gray]{0.95} \textit{Closed-source} & & & & & \\
GPT-Image-1.5$^{\dagger}$ \citep{gptimage15_model_card} & 8.391 & 8.890 & 8.577 & 8.148 & \textbf{9.070} \\
Nano Banana Pro \citep{google2025nanobanana} & 8.540 & 8.973 & 8.499 & 8.952 & 8.765 \\
Seedream5 lite \citep{seedream2025seedream}& 8.914 & 9.272 & 8.623 & \underline{9.525} & 8.880 \\
GPT-Image-2$^{\dagger}$ \citep{gptimage2_model_card} & \underline{9.111} & \underline{9.554} & \underline{8.823} & 9.478 & \underline{9.052} \\
\midrule
\textbf{\ours} & \textbf{9.171} & \textbf{9.619} & \textbf{8.955} & \textbf{9.620} & 8.883 \\
\bottomrule
\end{tabular}
}
\vspace{1mm}
\begin{minipage}{0.9\linewidth}
\scriptsize $^{\dagger}$ Due to platform restrictions, GPT-Image-1.5 failed to generate results for 168 test cases and GPT-Image-2 failed to generate results for 134 test cases. The reported metrics are calculated excluding these missing instances. Both GPT models were queried via the official API with the quality parameter set to \texttt{high}.
\end{minipage}
\end{table}

\paragraph{Multi-Garment Scenarios.}
The complexity of the try-on task escalates significantly in multi-garment scenarios, where models must coordinate multiple reference items, such as tops, bottoms, and various accessories.

A critical observation from Table \ref{tab:taobao_multi_results} is the \textbf{performance collapse of general-purpose image editing models} (e.g., FireRed-Image-Edit-1.1~\citep{firered2026rededit}, QwenEdit-2511~\citep{wu2025qwen}) when transitioning from single to multi-garment tasks. This degradation is primarily attributed to two bottlenecks. First, these models struggle with multi-garment coordination. They frequently omit specific garments or fail to resolve complex layering and occlusion relationships, leading to illogical garment combinations. Second, as the number of visual conditions increases, the task often exceeds the models' inherent capability boundaries, resulting in catastrophic generative failures where both the person's identity and the image's overall structure break down entirely.

In contrast, \ours maintains remarkable stability and achieves the highest overall results among all tested models. By leveraging our unified MMDiT architecture and specialized multi-garment training pipeline, \ours effectively manages the "stress test" of complex multi-piece outfits. It maintains high \textit{Garment Fidelity} for every individual item and preserves \textit{Physical and Structural Logic} without anatomical distortion. This demonstrates that \ours possesses the advanced visual reasoning required to handle industrial-grade multi-garment coordination, bridging the gap between experimental generation and professional-grade virtual fitting solutions.

\paragraph{Academic Benchmarks.}

While the results on the Tstars-VTON Benchmark unequivocally demonstrate our model's superiority in complex real-world scenarios, we further evaluate our framework on academic benchmarks to ensure comprehensive comparability with existing methods. Specifically, we conduct experiments on the two most widely recognized public datasets: VITON-HD~\citep{choi2021vitonhd} and DressCode~\citep{morelli2022dresscode}. 

VITON-HD is a high-resolution dataset specifically focused on upper-body garments, providing a standard testing set of 2,032 pairs. DressCode offers a more diverse and complex setting, featuring full-body person images categorized into upper-body, lower-body, and dresses, comprising 5,400 testing pairs.

In the literature, performance on these datasets is typically evaluated under two distinct protocols: paired and unpaired settings. The paired setting acts as a reconstruction-based assessment, where the model aims to synthesize the original garment back onto the person using a masked image. While this measures inpainting capability, it suffers from an inherent bias toward the ground-truth image and fails to represent real-world utility. Conversely, the unpaired setting requires the model to transfer a different reference garment onto the target person. Because this protocol more faithfully reflects the generalization required for practical virtual try-on applications, we predominantly focus our comparative analysis on the unpaired setting.

As shown in Table \ref{tab:unpaired_results}, \ours achieves state-of-the-art quantitative results against specialized academic baselines. Notably, our training dataset does not incorporate any data from either VITON-HD or DressCode. As a result, the superior performance observed on these benchmarks clearly demonstrates the strong zero-shot generalization and robust synthesis capabilities of our model when applied to entirely unseen data distributions.

\begin{table}[t]
\centering
\caption{\textbf{Quantitative comparison on VITON-HD and DressCode benchmarks under the unpaired setting.}}
\label{tab:unpaired_results}

\begin{tabular}{lcccc}
\toprule
\multirow{2}{*}{Method} & \multicolumn{2}{c}{VITON-HD} & \multicolumn{2}{c}{DressCode} \\
\cmidrule(lr){2-3} \cmidrule(lr){4-5}
 & FID $\downarrow$ & KID $\downarrow$ & FID $\downarrow$ & KID $ \downarrow$ \\
\midrule
Leffa \citep{zhou2024learning} & 10.446 & 2.640 & 20.099 & 13.506 \\
CatVTON \citep{chong2024catvtonconcatenationneedvirtual} & 10.552 & 2.272 & 5.872 & 1.606 \\
FitDiT \citep{jiang2024fitdit} & 9.979 & 1.478 & 4.805 & 0.712 \\
FastFit \citep{chong2025fastfitacceleratingmultireferencevirtual}& \underline{8.629} & \underline{0.665} & \textbf{4.397} & \underline{0.553} \\
\midrule
\textbf{\ours} & \textbf{8.485} & \textbf{0.528} & \underline{4.541} & \textbf{0.458} \\
\bottomrule
\end{tabular}
\end{table}

\begin{figure}
    \centering
    \includegraphics[width=\linewidth]{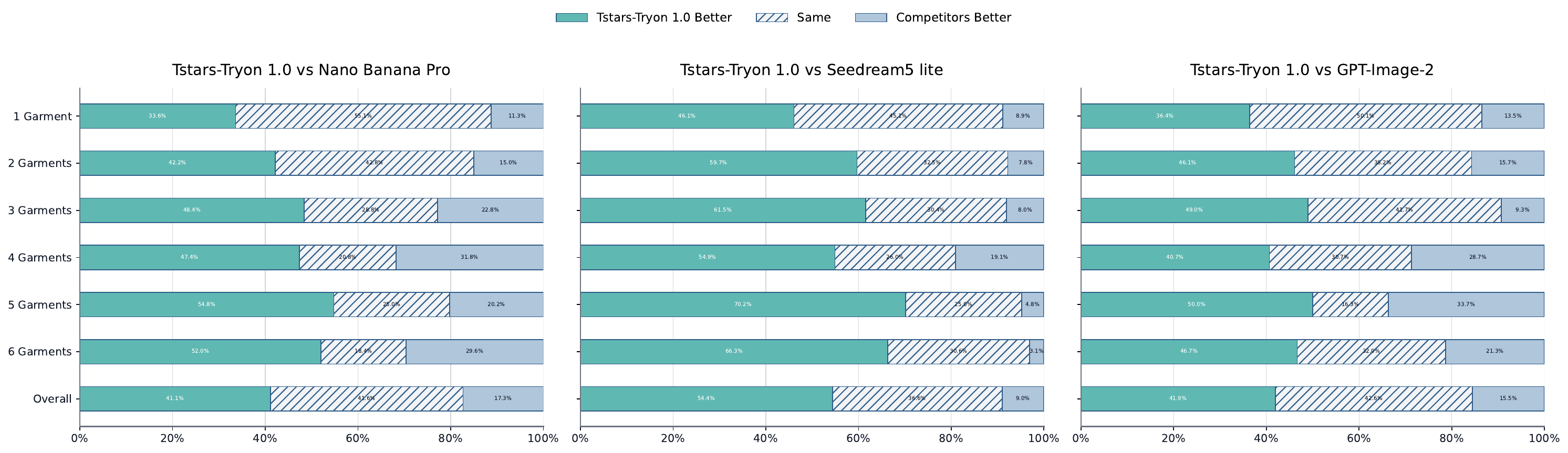}
    \caption{\textbf{Human evaluation comparison.} GSB Evaluation of \ours against Nano Banana Pro, Seedream5 Lite and GPT-Image-2 grouped by the number of reference garments. \ours consistently outperforms competitors overall, with its advantage becoming increasingly pronounced as the task complexity (number of garments) escalates.}
    \label{fig:gsb}
\end{figure}

\subsection{Human Evaluation}
While quantitative metrics provide a standardized measure of performance, they often fall short of capturing human perceptual preferences, which are the ultimate standard for commercial applications. To complement our quantitative analysis, we conducted a comprehensive human evaluation to assess the subjective visual quality of our generated images. 

We benchmarked \ours against three strong closed-source competitors identified in the quantitative analysis: Nano Banana Pro, Seedream5 lite, and GPT-Image-2. Evaluators were presented with randomized anonymized image pairs (ours vs. competitor) alongside the reference conditions and were asked to choose the better result or declare a tie (``Same''). The results are illustrated in Figure \ref{fig:gsb}.

\paragraph{Overall Superiority.} 
As shown in the ``Overall'' category, \ours demonstrates a definitive advantage over all three top-tier commercial models. When compared to Nano Banana Pro, \ours is preferred 41.1\% of the time, with 41.6\% rated as ties, and loses in only 17.3\% of cases. Performance against GPT-Image-2 follows a very similar pattern, where \ours achieves a 41.9\% win rate, 42.6\% ties, and only a 15.5\% loss rate. Against Seedream5 lite, our advantage is even more striking, achieving a 54.4\% win rate compared to a mere 9.0\% preference for the competitor. The high proportion of ``Same'' votes in the overall metrics indicates that while proprietary models can occasionally produce acceptable baseline results, \ours consistently pushes the upper bound of visual quality. Notably, although Seedream5 lite achieved slightly higher absolute scores than Nano Banana Pro in our quantitative metrics, it suffers a larger defeat margin here. This divergence occurs because GSB evaluation measures relative preference frequency rather than absolute score magnitude—a marginal visual advantage and a massive quality gap both register identically as a single ``Win'' in pairwise comparisons.

\paragraph{Robustness in High-Complexity Scenarios.} 
A compelling trend emerges when analyzing human preference across varying numbers of garments. In relatively simple single-garment try-on tasks, competitors manage to maintain a higher rate of ``Same'' evaluations. However, as the task complexity escalates to coordinate more garments simultaneously, the performance gap widens drastically.

For instance, against Nano Banana Pro, our win rate surges from 33.6\% (1 Garment) to a peak of 54.8\% (5 Garments). This upward trajectory is consistent across all competitors: against GPT-Image-2, our win rate increases from 36.4\% (1 Garment) to 50.0\% (5 Garments), and against Seedream5 lite, it jumps from 46.1\% (1 Garment) to an overwhelming 70.2\% (5 Garments). Correspondingly, the ``Same'' ratios drop significantly in these complex scenarios. This user study perfectly aligns with our quantitative findings: while existing models struggle to handle the intricate spatial layering, severe occlusions, and multi-condition conditioning required for full-outfit generation, \ours exhibits exceptional robustness and structural reasoning. This highly stable performance under extreme generative stress underscores the readiness of \ours for industrial deployment.

\subsection{Qualitative Results}
\begin{figure}
    \centering
    \includegraphics[width=1\linewidth]{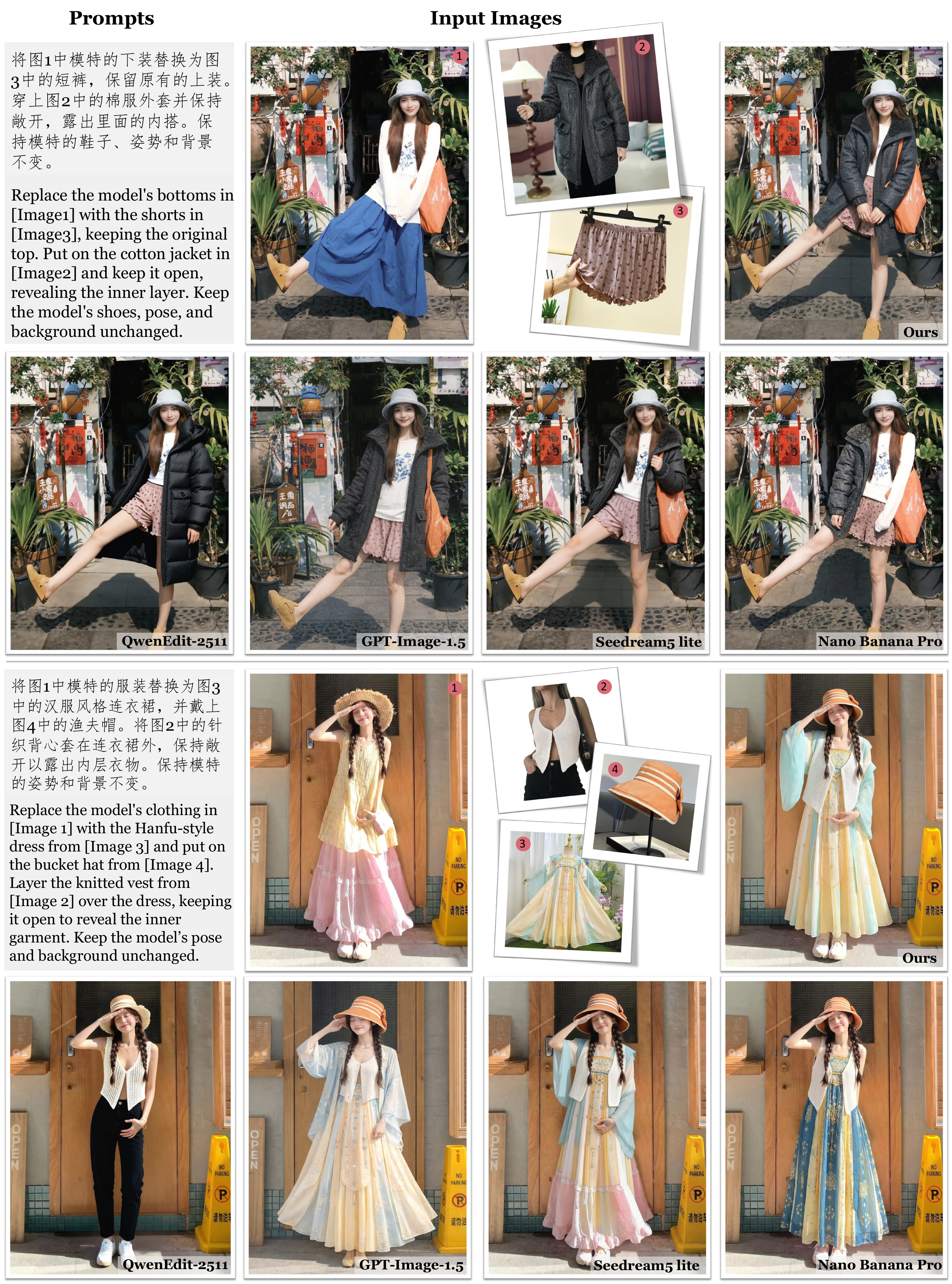}
    \caption{\textbf{Qualitative comparison of multi-garment and accessory try-on.} Compared to baseline models, \ours (Ours) more accurately follows text instructions and precisely reconstructs garment details.}
    \label{fig:comp1}
\end{figure}

\begin{figure}
    \centering
    \includegraphics[width=1\linewidth]{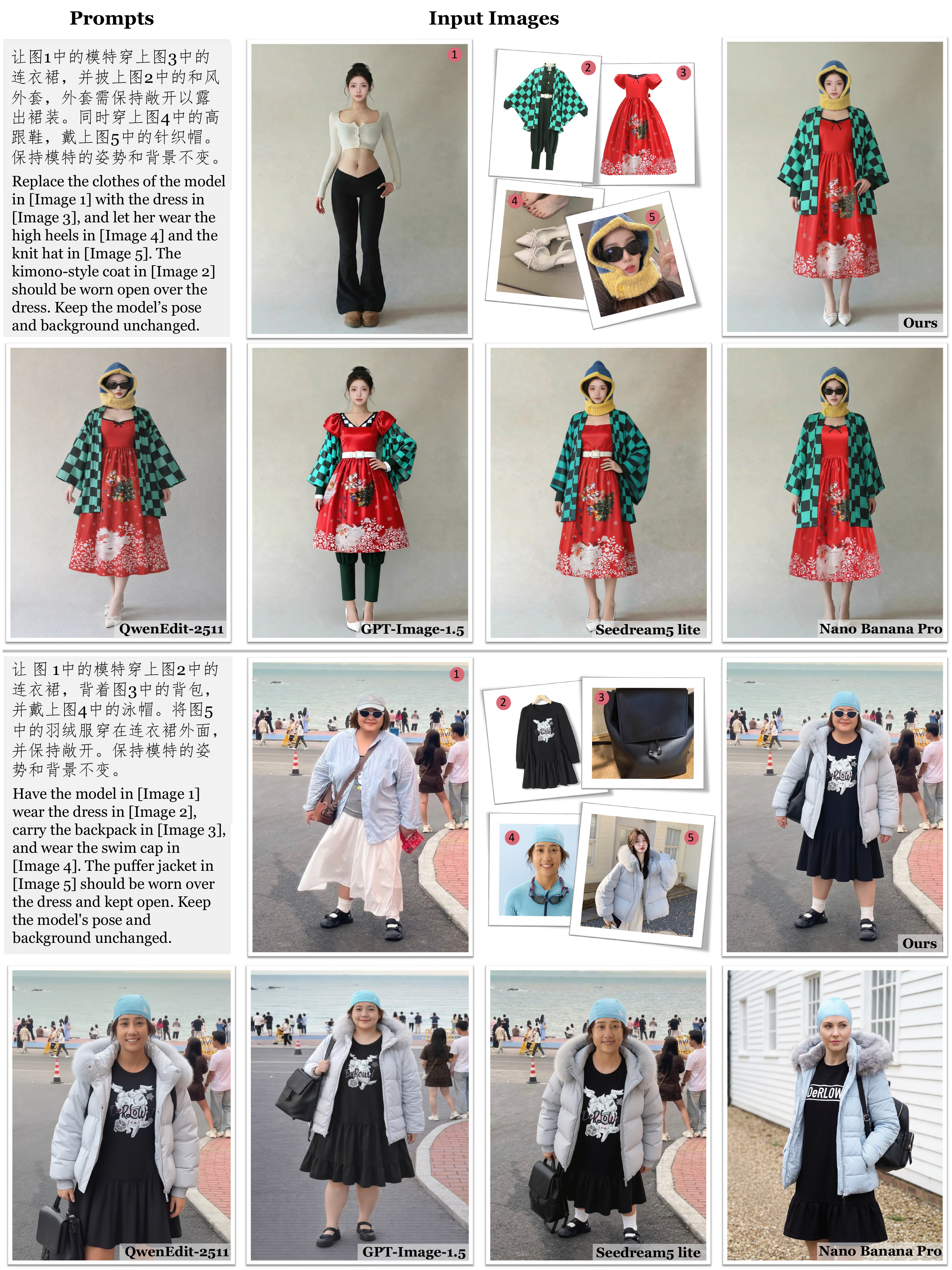}
    \caption{\textbf{Qualitative comparison under complex layered outfits and diverse human characteristics.} Our model demonstrates significant advantages in handling cross-style combinations and preserving complex backgrounds and identity.}
    \label{fig:comp2}
\end{figure}

\begin{figure}
    \centering
    \includegraphics[width=0.95\linewidth]{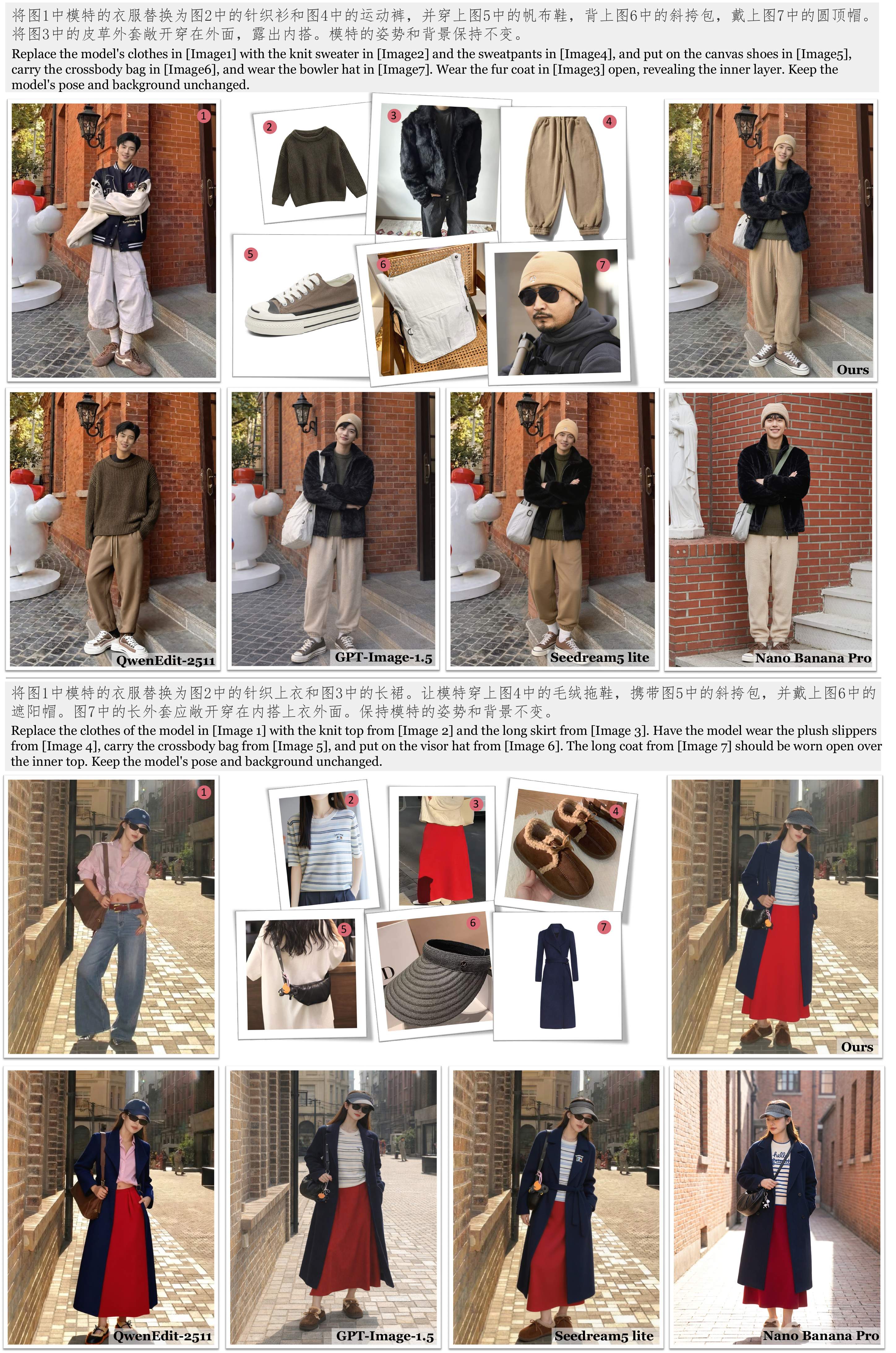}
    \caption{\textbf{Qualitative comparison of virtual try-on under extreme multi-condition scenarios (up to 6 garments).} When given a massive number of reference images, baselines suffer from item omission or identity degradation, whereas our model maintains high stability and semantic alignment.}
    \label{fig:comp3}
\end{figure}
To intuitively evaluate the performance of different models in complex multi-garment virtual try-on tasks, we present qualitative comparisons in Figures~\ref{fig:comp1}, ~\ref{fig:comp2}, and~\ref{fig:comp3}. In these scenarios, the models are required to rationally dress the source model with multiple reference images (covering tops, pants, skirts, dresses, coats, shoes, hats, and bags) based on complex textual prompts, while strictly maintaining the original pose and background. Experimental results demonstrate that our proposed method significantly outperforms existing top open-source or proprietary models across three key dimensions:
\begin{itemize}[leftmargin=*]
    \item \textbf{Extreme Robustness: Stable Identity, Pose, and Background Consistency.} A core challenge in try-on tasks is redrawing garments without disrupting the unedited regions of the source image. In Figure~\ref{fig:comp2} (bottom) and Figure~\ref{fig:comp3} (top), Nano Banana Pro, GPT-Image-1.5, and Seedream5 lite experience severe "Identity Degradation" during complex full-body replacements—drastically altering the model's facial shape and body type, or hallucinating entirely incorrect backgrounds. Additionally, in Figure~\ref{fig:comp1} (top), all baseline models fail to preserve the model's original hand pose. Our method, through a more precise local control mechanism, flawlessly retains the model's original facial features, body posture, and intricate natural/street backgrounds while achieving complex full-body garment transfers.
    \item \textbf{High Realism: Exceptional Fidelity of Garment Details.} Observations from the generated images indicate that whether dealing with complex patterns (e.g., the kimono checkered pattern in Figure~\ref{fig:comp2} (top)), specific materials (e.g., the fur coat in Figure~\ref{fig:comp3} (top) and plush slippers in Figure~\ref{fig:comp3} (bottom)), or non-standard accessories (e.g., the swim cap in Figure~\ref{fig:comp2} (bottom)), our model highly restores the texture and structural features of the reference images. Conversely, baseline models GPT-Image-1.5 and Nano Banana Pro hallucinate the color of the bottoms (Figure~\ref{fig:comp3} (top)), and Seedream5 lite generates a logo inconsistent with the reference coat (Figure~\ref{fig:comp3} (bottom)). Furthermore, as shown in Figure~\ref{fig:comp1} (bottom), both Seedream5 lite and Nano Banana Pro severely confuse the color and style of the skirt.
    \item \textbf{Unprecedented Flexibility: Superior Instruction Following and Multi-Garments Generation.} As the number of reference garments increases (e.g., combinations of up to 6 items as shown in Figure~\ref{fig:comp3}), existing open-source and closed-source models generally suffer from severe "feature omission" or "semantic confusion." For instance, in Figure~\ref{fig:comp3} (top), QwenEdit-2511 misses the crossbody bag, while GPT-Image-1.5 and Nano Banana Pro confuse the color of the bottoms. In contrast, our model demonstrates strong multimodal contextual understanding. It not only accurately places all reference items on the correct body parts but also precisely executes complex spatial logic instructions such as "keep open, revealing the inner layer" (see Figures~\ref{fig:comp1} and ~\ref{fig:comp2}).
\end{itemize}

In summary, the qualitative analysis validates that our model exhibits industry-leading flexibility, realism, and robustness in generation quality when handling highly complex, real-world "multi-garment try-on" and "multi-condition combination" scenarios.

%% file: Sections/demonstrations.tex
\section{Demonstrations}

\subsection{Single-Garment}
\begin{figure}[htbp]
    \centering
    \includegraphics[width=1\linewidth]{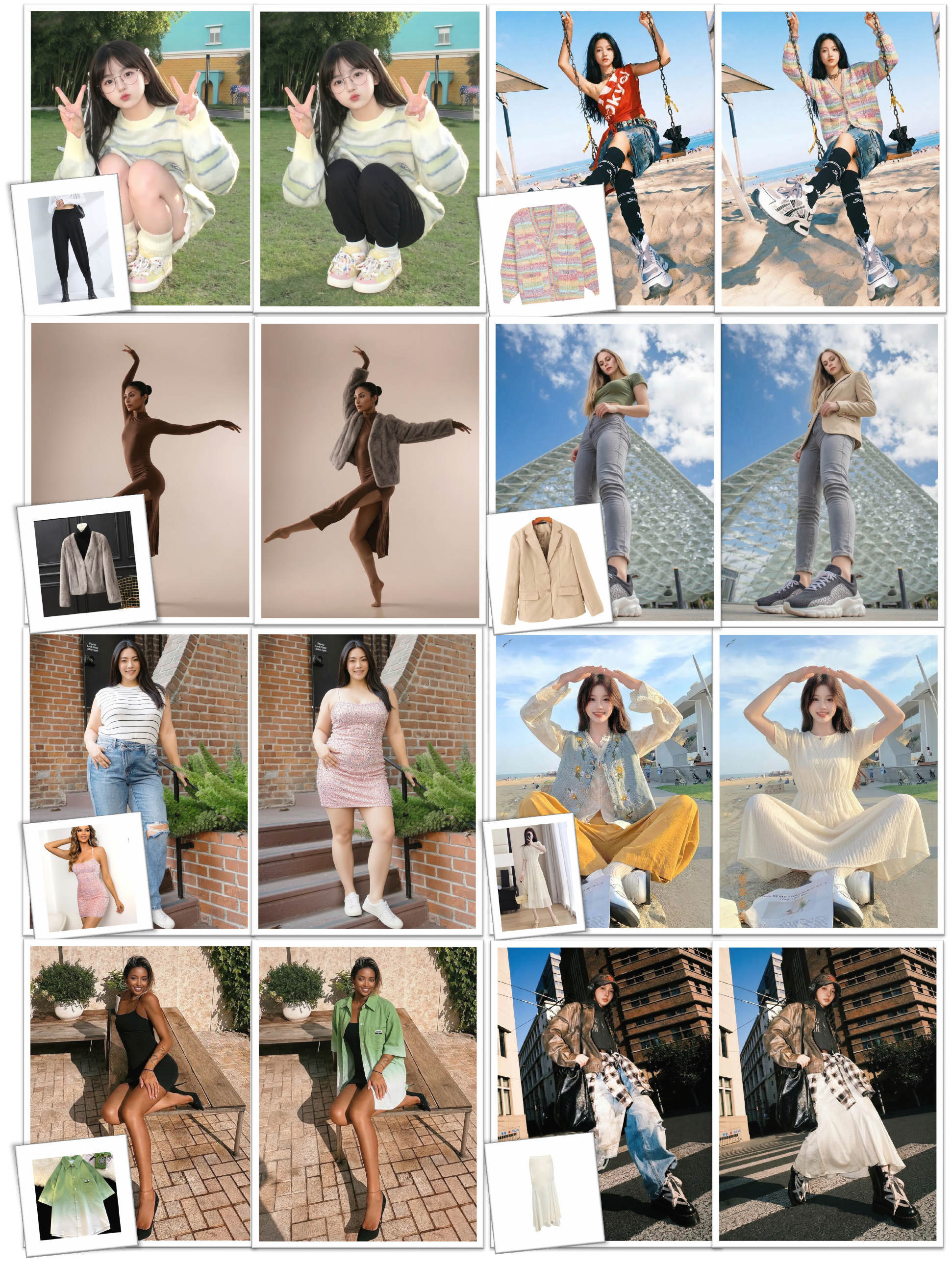}
    \caption{\textbf{Qualitative demonstrations of single-garment try-on.} Showcasing \ours extreme robustness, precise preservation capabilities (identity, pose, background, body shape), and high-fidelity rendering of complex materials across varying perspectives and input conditions.}
    \label{fig:qual_single}
\end{figure}
This Figure~\ref{fig:qual_single} showcases the extreme robustness and high-fidelity rendering capabilities of \ours across a variety of challenging scenarios. In the top row, the system demonstrates its ability to map 2D flat-lay garments—such as black trousers (left) and a colorful knit cardigan (right)—onto complex human poses, including crouching and swinging, while maintaining accurate material textures and lighting. The second row highlights spatial adaptation and material realism, successfully fitting a fuzzy grey jacket onto a ballet dancer in motion (left) and a tailored blazer onto a subject in a sharp low-angle perspective (right) without geometric distortion. The third row emphasizes the model’s strict adherence to body shape and complex occlusions, accurately rendering a patterned pink dress on a plus-size model (left) and a flowing cream dress on a seated subject (right), ensuring realistic folds and shadows. Finally, the bottom row illustrates the meticulous preservation of identity and background; the system seamlessly integrates a green ombre shirt (left) and a long white skirt (right) into scenes with intricate outdoor lighting and architectural backgrounds, keeping the user’s facial features and the environment’s structural integrity perfectly intact.
\subsection{Multi-Garment}
\begin{figure}[htbp]
    \centering
    \includegraphics[width=1\linewidth]{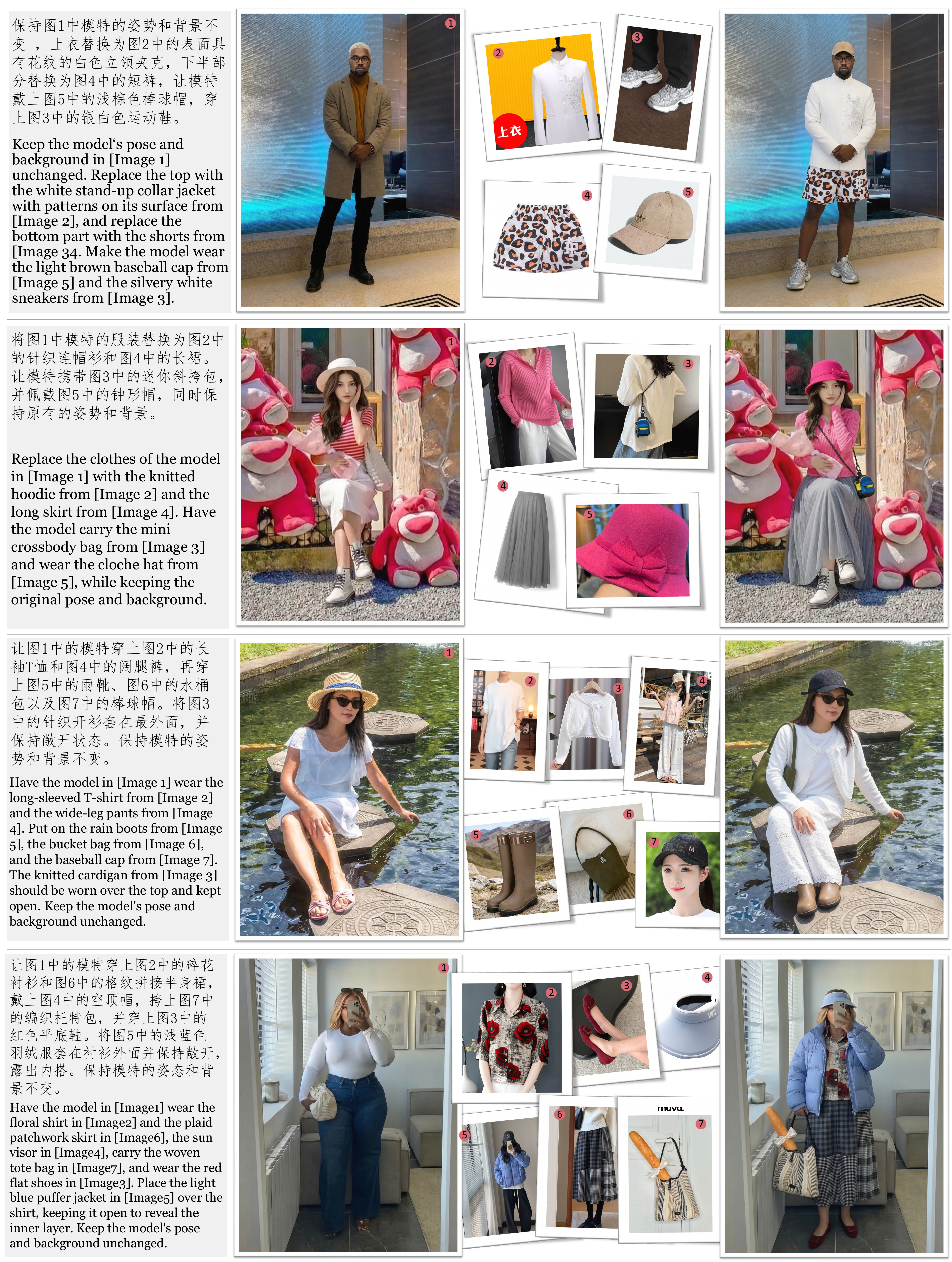}
    \caption{\textbf{Demonstrations of multi-garment try-on outfit composition.} Highlighting the model's capability for reasonable multi-garment layering, diverse accessory try-on, and strict preservation of user attributes including diverse body types.}
    \label{fig:qual_multi}
\end{figure}
This Figure~\ref{fig:qual_multi} showcases the advanced capabilities of the system to generate complex, multi-item outfits across various users and environments. In the top row, the system demonstrates precise multi-item mapping for a male model in an upscale indoor setting, seamlessly integrating an intricate white stand-up collar jacket and patterned leopard shorts with multiple accessories like a cap and sneakers, showcasing robust cross-item geometric alignment. The second row features a complete style overhaul for a female model posed among numerous stuffed animals; her simple original outfit is replaced with a coordinated pink knitted hoodie and grey pleated skirt, accented by a new cloche hat and a crossbody bag. The model retains her pose and the complex, occluding background, demonstrating high-fidelity details in a dense environment. In the third row, the focus is on reasonable multi-garment layering in an outdoor setting. The model accurately layers a long-sleeved T-shirt and wide-leg pants under an open white knitted cardigan, while adding multiple accessories like a bucket bag and a different hat, ensuring realistic drape and interactions with the water background. Finally, the bottom row underscores the model’s ability to strictly preserve user attributes, specifically for a plus-size body type. A complex ensemble, including a floral shirt, a patterned skirt, and an open light blue puffer jacket, is accurately fitted without distortion or artificial slimming, demonstrating effective layering of distinct textures and patterns.


\begin{figure}[h]
    \centering
    \includegraphics[width=1\linewidth]{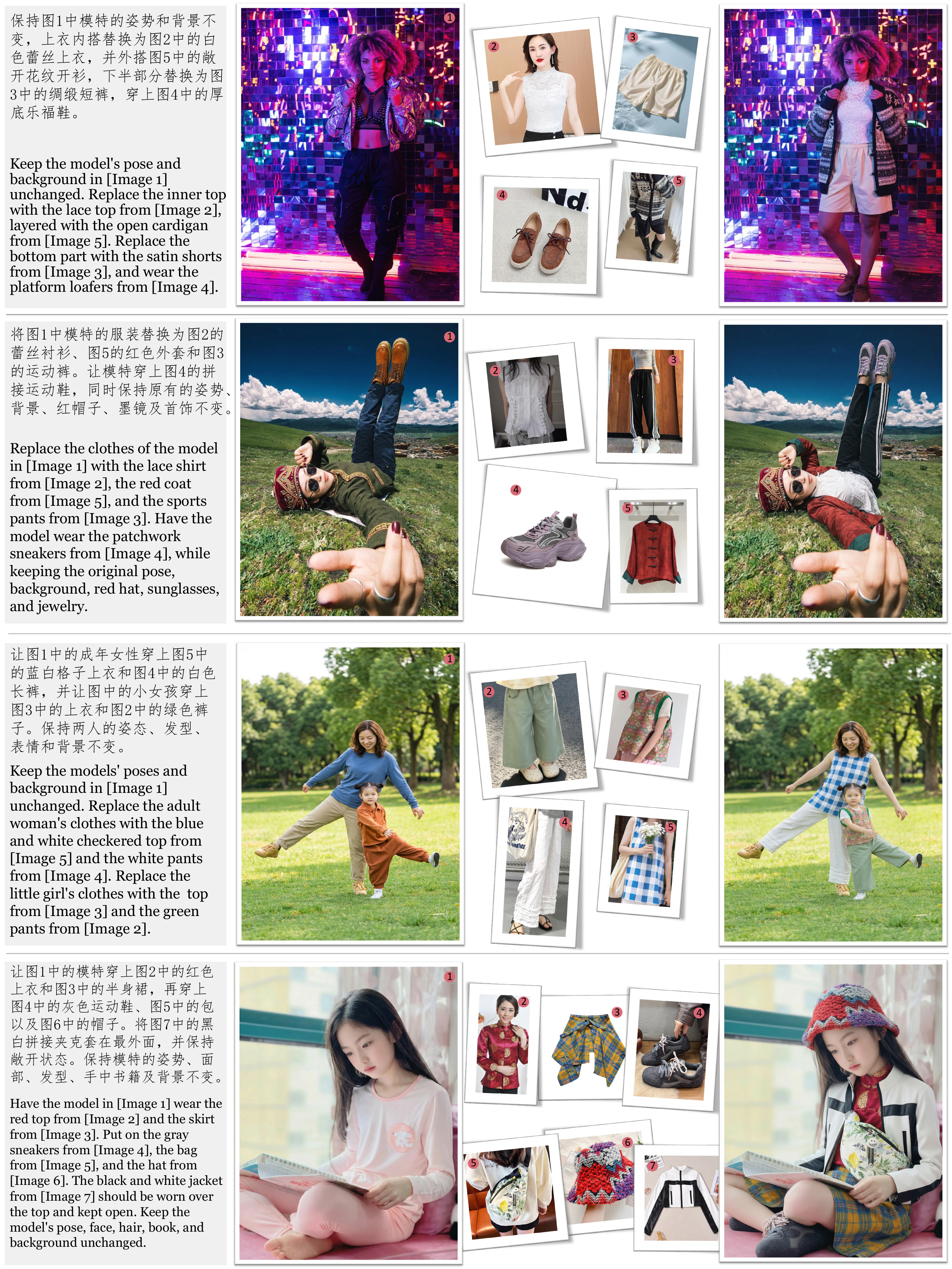}
    \caption{\textbf{Versatile multi-item synthesis.} Advanced applications in multi-item synthesis under heterogeneous lighting, unconventional perspectives, and multi-subject interactions}
    \label{fig:showcase_multi}
\end{figure}
This Figure~\ref{fig:showcase_multi} highlights the model's versatility in handling unconventional orientations and intricate lighting interplay. In the first row, the system excels in low-light, high-contrast neon environments, accurately rendering the delicate lace of the inner top and the sheen of satin shorts while maintaining consistent global illumination. The second row showcases extreme geometric robustness, where the model successfully adapts a red coat and sneakers to a challenging lying-down perspective with significant foreshortening. The third row demonstrates a breakthrough in simultaneous multi-subject synthesis, flawlessly executing garment swaps for both an adult and a child within a shared natural setting while preserving their interactive poses and distinct body scales. Finally, the bottom row illustrates the handling of complex item ensembles and occlusion; a hat, bag, and layered jacket are realistically integrated onto a seated model holding a book, achieving exceptional perceptual consistency and physical detail across a dense combination of accessories.

\begin{figure}[t]
    \centering
    \includegraphics[width=1\linewidth]{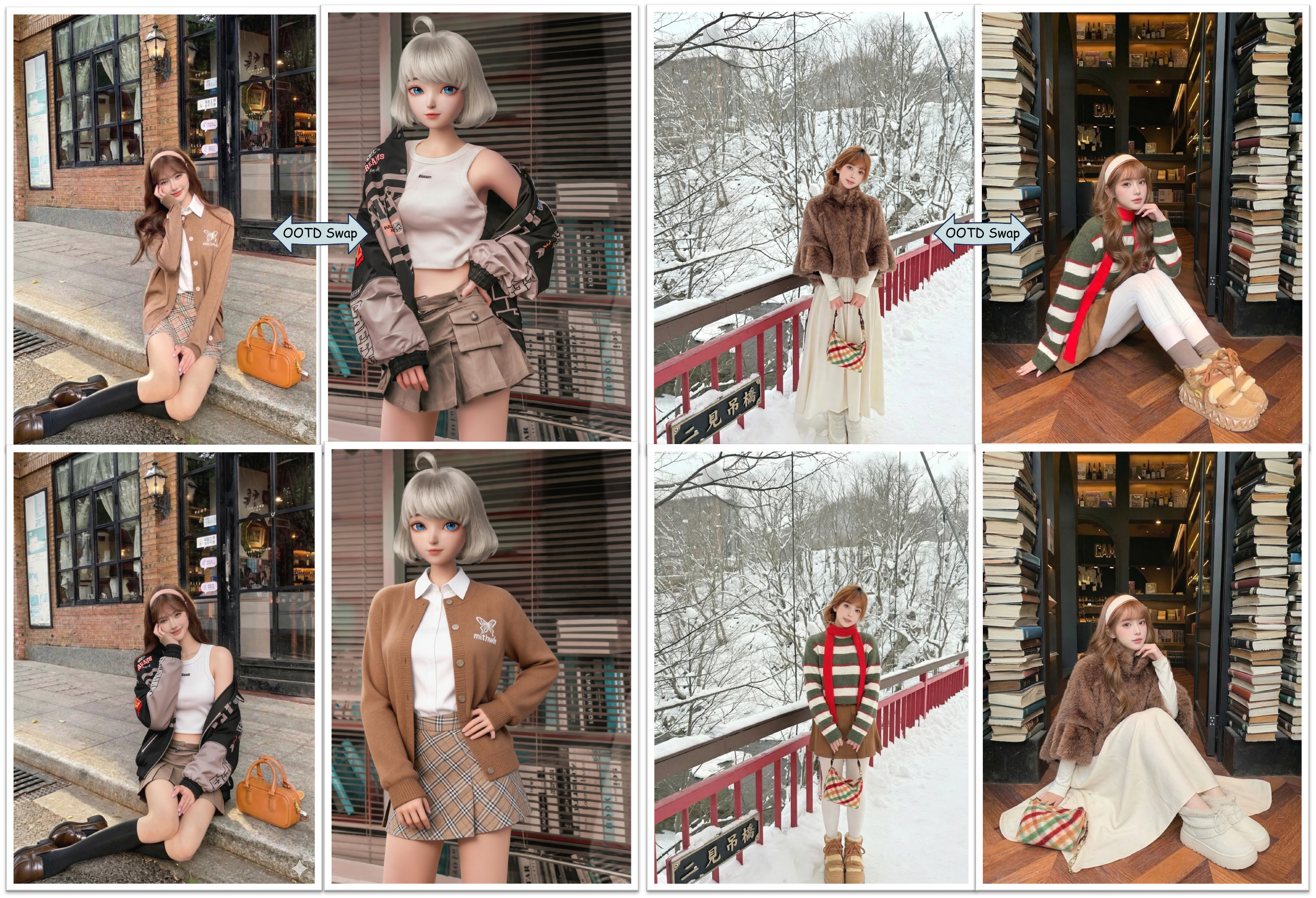}
    \caption{\textbf{Demonstrations of holistic OOTD (Outfit of the Day) swapping across diverse subjects, poses, and domains.} \ours flawlessly transfers entire ensembles between different individuals, including cross-domain transfers between real humans and 3D avatars, while strictly preserving identities and backgrounds.}
    \label{fig:showcase_ootd}
    \vspace{-4mm}
\end{figure}

Figure~\ref{fig:showcase_ootd} illustrates a highly flexible application of \ours: the holistic ``OOTD (Outfit of the Day) Swap.'' This task requires the model to extract and transfer a complete stylistic ensemble from a single reference image—encompassing tops, bottoms, outerwear, footwear, and accessories, etc.—between two entirely different subjects. The model effortlessly achieves this mutual exchange while seamlessly adapting the garments to different body types, challenging postures (e.g., adapting an outfit from a standing pose in the snow to a seated pose in a library), and varying environmental lighting conditions. Notably, the left panel demonstrates exceptional cross-domain robustness, successfully swapping a complex real-world outfit with a 3D animated character. \ours perfectly fits the physical garments onto the stylized 3D geometry without compromising the character's unique aesthetic or the human subject's photorealism. Throughout these massive holistic transformations, the user identity characteristics and highly complex background contexts are strictly preserved, underscoring the model's superior capability in disentangling clothing semantics from spatial and identity features.

\subsection{Semantic expansion}
\begin{figure}[htb]
    \centering
    \includegraphics[width=1\linewidth]{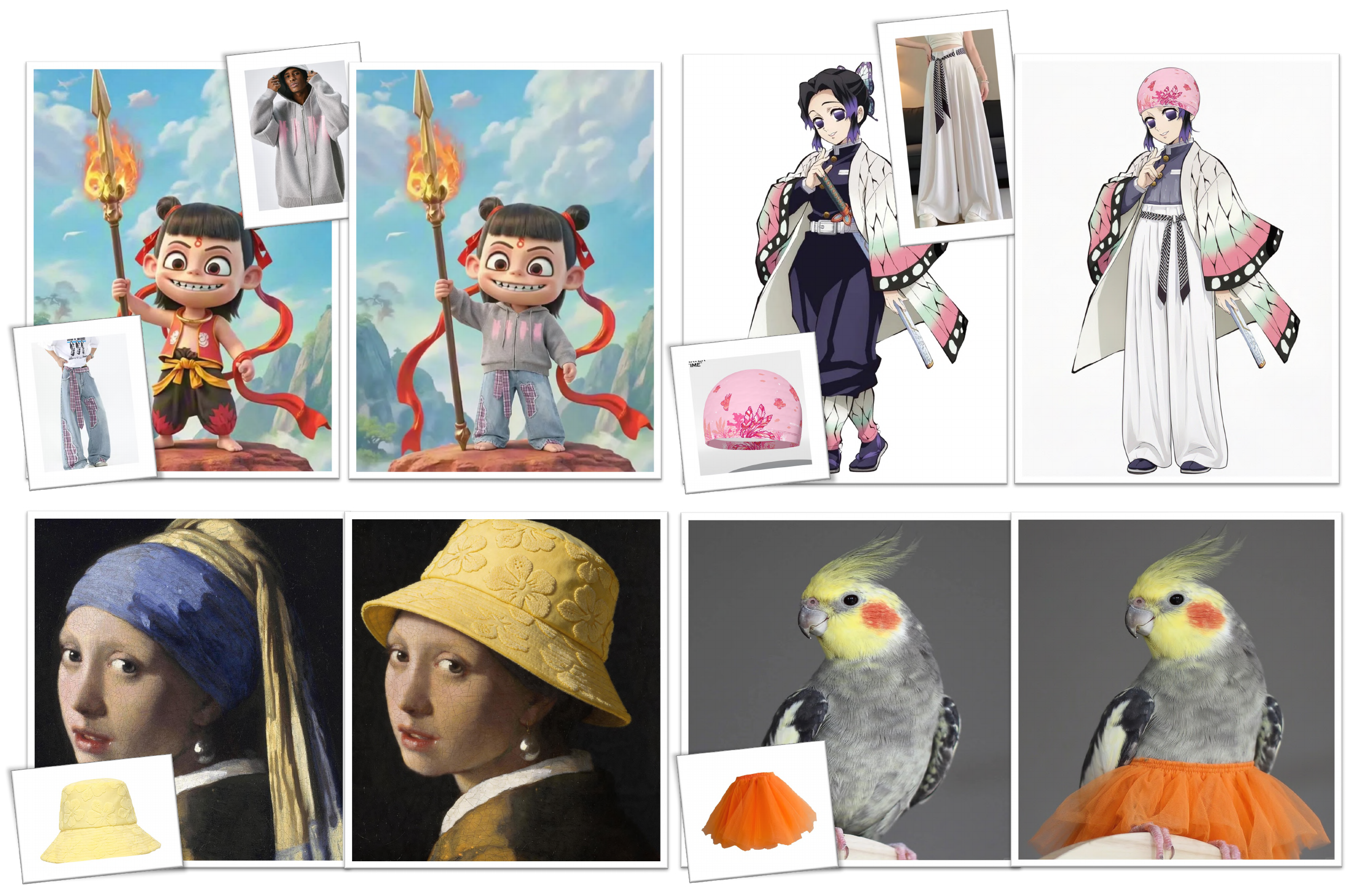}
    \caption{\textbf{Cross-domain virtual try-on capabilities.} Showcasing the model's flexible semantic extensibility across diverse non-photorealistic styles and non-human subjects.}
    \label{fig:semantic_generalization}
\end{figure}
To demonstrate the powerful semantic generalization capabilities of \ours, Figure~\ref{fig:semantic_generalization} illustrates its ability to transcend the boundaries of standard human photography and adapt to highly unconventional, out-of-domain subjects. Moving from top-left to bottom-right, the model seamlessly applies modern streetwear (a zip-up hoodie and jeans) onto a 3D animated character(Row 1, left), accurately handling highly stylized, non-standard body proportions without spatial distortion. It also successfully adapts to the flat geometric domain of 2D anime, naturally fitting a patterned hat and wide-leg pants onto an illustrated character while respecting the original artistic aesthetic(Row 1, right). Furthermore, the system exhibits extraordinary cross-modality blending by replacing the iconic headwear in a classical oil painting ("Girl with a Pearl Earring") with a modern floral bucket hat, remarkably matching the historical lighting, shadows, and brushstroke textures of the original artwork(Row 2, left). Finally, the model's structural understanding extends even to non-anthropomorphic subjects, flawlessly rendering an orange tutu skirt onto a bird(Row 2, right). These diverse cross-domain applications prove that \ours has learned deep, generalizable semantic representations of garments and spatial relationships, rather than merely overfitting to real-world human pose priors.

%% file: Sections/deployment.tex
\section{Industrial-Scale Deployment}

\begin{figure}
  \centering
  \makebox[\linewidth][c]{%
    \includegraphics[width=1.05\linewidth]{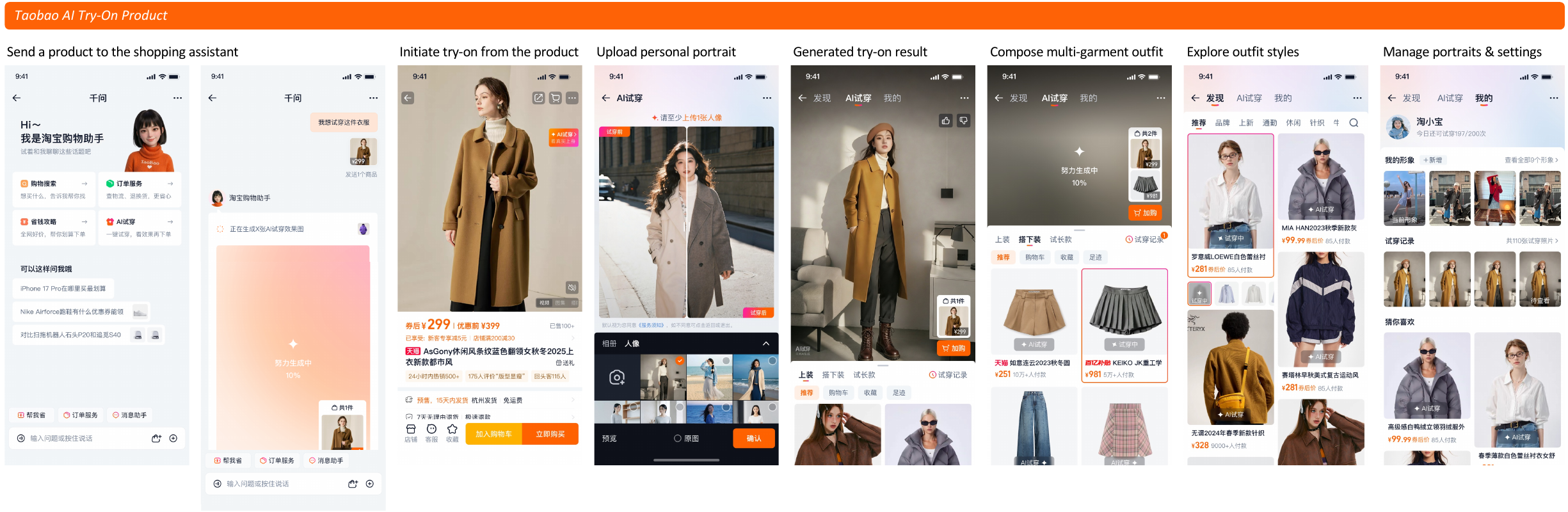}%
  }
  \caption{\textbf{Industrial application.} Industrial-scale deployment of Tstars-Tryon as the "AI Try-On" service on the Taobao App, illustrating the complete consumer-facing user journey from try-on initiation, portrait upload, single-/multi-garment try-on generation, to outfit-style exploration and personal portrait management.}
  \label{fig:product}
\end{figure}

The proposed algorithm has been fully deployed within the Taobao App and is publicly accessible to end consumers as the "AI Try-On" service. As illustrated in Figure~\ref{fig:product}, the system supports a complete consumer-facing user journey: users can initiate a try-on request either through the in-app shopping assistant or directly from a product detail page, and, after uploading a personal portrait, can try on a wide range of apparel items as well as freely compose DIY multi-garment outfits.

To the best of our knowledge, this constitutes one of the largest production-scale deployments of virtual try-on technology to date. The system has served several million users and fulfilled tens of millions of try-on requests, demonstrating its robustness, scalability, and commercial viability under real-world industrial workloads. We further plan to roll out the service to the entire Taobao user base, where the system is expected to handle tens of millions of try-on requests per day, driving the large-scale adoption of virtual try-on technology in e-commerce scenarios. This large-scale deployment further validates that the proposed approach effectively resolves the long-standing trade-off between C-end serving cost and generation quality, enabling virtual try-on to transition from a research prototype to a fully commercialized consumer-facing product.

%% file: Sections/acknowledgments.tex
\section{Acknowledgments}

All contributors are listed in alphabetical order by their last names.

\noindent\textbf{Engineering Contributors: } Yichao Cai, Donglai Ge, Zhiwei Han, Shuaiqi Jia, Tao Lan, Jiacheng Li, Yi Li, Kan Liu, Xu Liu, Zhenxiao Liu, Weiyi Lu, Chong Ma, Lin Qu, Chuanli Wang, Daisong Wang, Hanlun Wang, Lujie Wang, Yi Wang, Xinjiang Wu, Jiawei Zhang, Lei Zhang, Mao Zhou, Guoxuan Zhu, Tianfu Zhu, Yongjie Zhu

\noindent\textbf{Productization Contributors:} Yajun Bai, Jian Ding, Zhengni Guan, Mengli Huang, Nan Liu, Yating Sheng, Xudong Wu, Xiaoyu Zhu